\documentclass[10pt,twocolumn,letterpaper]{article}

\usepackage[pagenumbers]{cvpr} 

\usepackage{mathtools}
\usepackage{multirow}
\usepackage{makecell}
\usepackage{enumitem}
\usepackage{algorithm}
\usepackage{algpseudocode}
\usepackage{wrapfig}
\usepackage{colortbl}
\usepackage{capt-of}
\usepackage{booktabs}
\usepackage{tabularx}

\newcommand{\method}{ALG\xspace}

\makeatletter

\newcommand{\remotefootnotetext}[2]{%
  \refstepcounter{footnote}%
  \edef\@thefnmark{\thefootnote}
  \label{#1-fn}%
  \insert\footins{%
    \reset@font\footnotesize
    \interlinepenalty\interfootnotelinepenalty
    \splittopskip\footnotesep
    \splitmaxdepth \dp\strutbox
    \floatingpenalty \@MM
    \hsize\columnwidth
    \@parboxrestore
    \hypertarget{#1-text}{}%
    \@makefntext{#2}%
  }%
}

\makeatother

\definecolor{cvprblue}{rgb}{0.21,0.49,0.74}
\usepackage[pagebackref,breaklinks,colorlinks,allcolors=cvprblue]{hyperref}

\def\paperID{35852} 
\def\confName{CVPR}
\def\confYear{2026}

\title{Improving Motion in Image-to-Video Models via Adaptive Low-Pass Guidance}

\author{June Suk Choi \quad Kyungmin Lee \quad Sihyun Yu \quad Yisol Choi\\
Jinwoo Shin \quad Kimin Lee\\
KAIST\\
{\tt\small \{w\_choi, kyungmnlee, sihyun.yu, yisol.choi, jinwoos, kiminlee\}@kaist.ac.kr}
}

\begin{document}


\maketitle
\begin{abstract}
Recent text-to-video (T2V) models have demonstrated strong capabilities in producing high-quality, dynamic videos.
To improve the visual controllability, recent works have considered fine-tuning pre-trained T2V models to support image-to-video (I2V) generation. 
However, such adaptation frequently suppresses motion dynamics of generated outputs, resulting in more static videos compared to their T2V counterparts.
In this work, we analyze this phenomenon and identify that it stems from the premature exposure to high-frequency details in the input image, which biases the sampling process toward a shortcut trajectory that overfits to the static appearance of the reference image. 
To address this, we propose adaptive low-pass guidance (\method), a simple training-free fix to the I2V model sampling procedure to generate more dynamic videos without compromising per-frame image quality.
Specifically, \method adaptively modulates the frequency content of the conditioning image by applying a low-pass filter at the early stage of denoising.
Extensive experiments show \method significantly improves the temporal dynamics of generated videos, 
while preserving or even improving image fidelity and text alignment. 
For instance, on the VBench test suite, \method achieves a 33\% average improvement across models in dynamic degree while maintaining the original video quality.
For additional visualizations and source code, see the \href{https://choi403.github.io/ALG/}{project page}.
\end{abstract}    
\section{Introduction} \label{sec:introduction}

 Generative models based on iterative denoising processes---such as diffusion \citep{ddpm, diffusion, scorebased, dhariwal2021diffusion} and flow-matching \citep{flowmatching, rectifiedflow} models---have emerged as a scalable framework for generating a variety of data, including images~\citep{glide, imagen, ldm, sdxl, sd3, li2024hunyuanditpowerfulmultiresolutiondiffusion}, videos~\citep{makeavideo, yu2023video, svd, gupta2024photorealistic, he2023latentvideodiffusionmodels, yu2025malt}, and audio~\citep{diffwave, makeanaudio}. In particular, they have enabled the challenging task of text-to-video (T2V) generation~\citep{villegas2022phenakivariablelengthvideo, ho2022imagenvideohighdefinition, lin2024opensoraplanopensourcelarge, veo, hong2022cogvideolargescalepretrainingtexttovideo, nvidia2025cosmosworldfoundationmodel} to synthesize temporally coherent, aesthetic, and diverse video sequences based on complex text prompt inputs.

However, T2V diffusion models often lack \emph{controllability}---for example, the ability to animate a specific input image or ground video content in existing visual concepts. 
To address this, recent works have explored image-to-video (I2V) generation models~\citep{i2vgenxl, svd, guo2023animatediff, xing2024dynamicrafter, ai2025magi1autoregressivevideogeneration, shi2024motion, ni2023conditional}, which generate videos conditioned on reference images. 
These models are typically built by fine-tuning large-scale T2V models~\citep{wan, hunyuan, ltx} to incorporate image and text inputs.
This approach has demonstrated promising results in generating high-quality, consistent videos from reference images.

\begin{figure*}[t]
  \centering
  \includegraphics[width=0.98\textwidth]{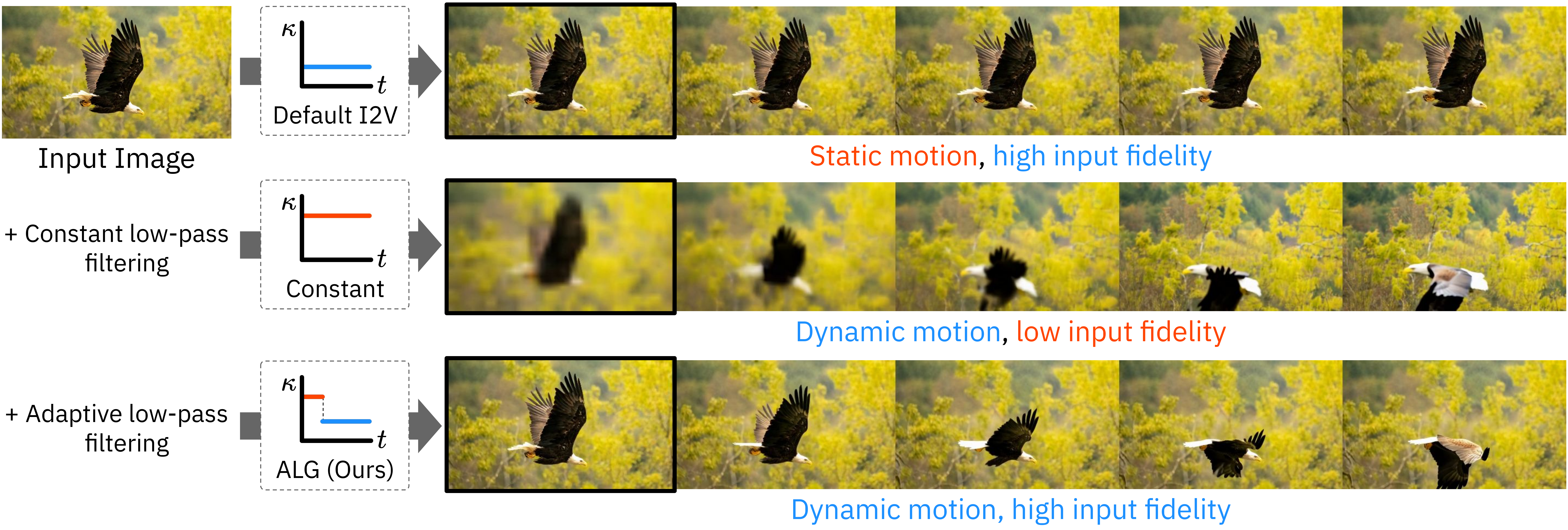}
  \vspace{-5pt}
  \caption{
  {\bf Overcoming suppressed motion dynamics of I2V models with \method.}
  I2V models achieve high image fidelity to the conditioning image, but they often fail to generate dynamic videos (first row). 
  We refer to this issue as \emph{suppressed motion dynamics}, which is due to the high-frequency details present in the reference image.
  As a simple fix, applying low-pass filter to the input image improves the motion dynamics, yet degrades the per-frame image quality and fidelity (second row).
  Our method, \method, applies low-pass filter to the conditioning image only at earlier steps, significantly enhancing the dynamic degree while preserving the image quality (third row).
  }
  \vspace{-0.1in}
  \label{fig:intro_qualitative}
\end{figure*}

Despite these advances, I2V models built on fine-tuned T2V architectures frequently produce much more static videos compared to their T2V counterparts, often adhering too closely to the reference image~\citep{svd, identifying, wu2024freeinitbridginginitializationgap} (Fig.~\ref{fig:intro_qualitative}, first row).
We hypothesize that this motion suppression is caused by the high-frequency components of the reference image, causing I2V models to lock onto these fine details during the generation process, preventing large-scale, coarse motions from developing. To test this, we start our analysis by quantifying the motion suppression effect in I2V models. 
We use models where both pre-trained T2V and fine-tuned I2V checkpoints are available, as a clean testbed to isolate the effects of the I2V conditioning mechanism while fixing other factors.
In our experiments, I2V models indeed generate more static videos (\emph{i.e.}, lower dynamic degree) compared to T2V models, even when they share the initial condition with T2V models (see Tab.~\ref{tab:t2v_to_i2v_suppression} for more details).

\begin{figure*}[t]
  \centering
  \includegraphics[width=0.98\textwidth]    
{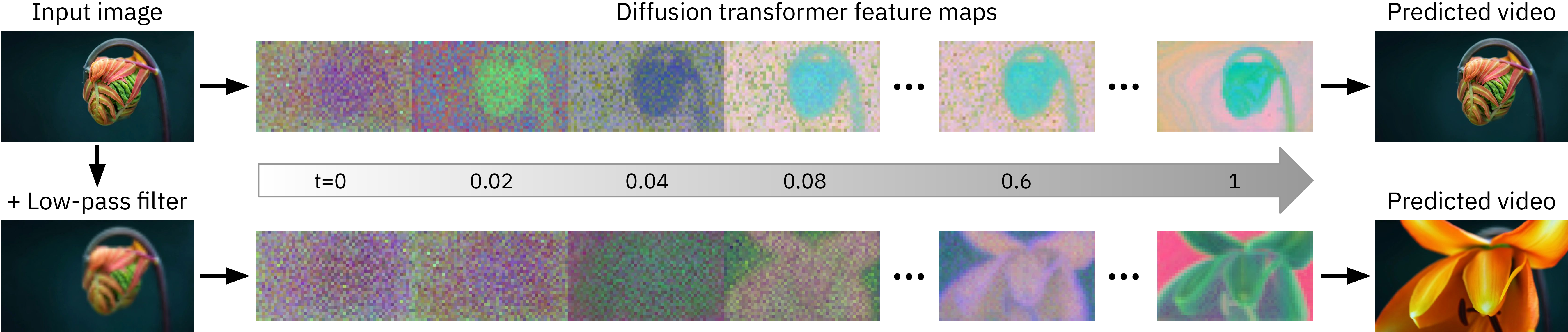}  
  \vspace{-0.05in}
  \caption{\textbf{
  Visualization of shortcut effect in I2V generation.}
  Intermediate feature map visualization from Wan 2.1~\citep{wan} reveal that default I2V generation (top) exhibits a ``shortcut'' completion where fine-grained details in the image appears quickly (yellow dashed box), which confines the trajectory and prevents coarse structure from forming, ending up with a static video.
  Applying a low-pass filter (bottom) suppresses this shortcut to allow details to emerge gradually, and such flexible trajectory helps generating dynamic motion.}

  \label{fig:shortcut}
  \vspace{-0.15in}
\end{figure*}

Next, we apply low-pass filter to input images to remove the high-frequency components. 
As shown in the second row of Fig.~\ref{fig:intro_qualitative}, the videos have more vivid and dynamic motion, though it loses fidelity from filtering.
This suggests that high-frequency details substantially influence motion.
Finally, we inspect the internal representation of the I2V model's denoiser backbone to further investigate model behavior regarding our hypothesis.
Fig.~\ref{fig:shortcut} visualizes how I2V generation process falls into a {\em shortcut} early in the trajectory, prematurely locking in fine details, which prevents large motions from evolving.
Low-pass filtering the input image mitigates this shortcut by allowing more flexibility in the later generation steps, leading to more dynamic videos.

Based on these observations, we introduce \emph{\textbf{A}daptive \textbf{L}ow-Pass \textbf{G}uidance} (\method), a simple training-free modification to the I2V model sampling procedure that significantly improves motion dynamics while maintaining video quality. The key idea of \method is \emph{adaptive conditioning} across timesteps during sampling. Specifically, we condition the model on a low-pass filtered version of the reference frame at early timesteps, and switch to the original reference frame at later timesteps. This simple fix effectively prevents the sampling trajectory from converging to a ``shortcut solution'' and enhances motion quality, while preserving fine-grained image details by reintroducing high-frequency information in later stages.
As a result, we observe that our solution achieves improved motion quality while maintaining video quality (see the third row in Fig.~\ref{fig:intro_qualitative}).

We extensively validate the effectiveness of \method on various recent open-source I2V models, including Wan~2.1, Wan 2.2~\citep{wan}, and LTX-Video~\citep{ltx}, using various benchmark datasets. For instance, we demonstrate that \method achieves an average of 33\% improvement of dynamic degree across various models in the VBench~\citep{vbench} I2V test suite while maintaining or even often improving video quality and input image fidelity, with no additional training.

\noindent In summary, our contribution is given as follows:
\begin{itemize}[leftmargin=*,itemsep=1mm]
    \item We identify the suppression of motion in I2V models and show low-pass filtering the initial frame mitigates this issue, but at the cost of reduced image fidelity and quality.
    \item We propose \method, a training-free, simple inference technique to enhance dynamic degree of I2V models by adaptively low-pass filtering the conditioning image.
    \item Experiments validate \method consistently enhances motion dynamics of I2V models (\emph{e.g.}, 33\% improvement on VBench), without losing image fidelity or video quality.
\end{itemize}

\section{Background} \label{sec:preliminaries}
\vspace{0.02in}\noindent
{\bf Flow matching.}~
Generative models based on the \emph{Flow Matching} (FM)~\citep{flowmatching, normalizingflow} learn to transform samples from a prior distribution $p_0 (\mathbf{x}) \coloneqq \mathcal{N}(\boldsymbol{0}, \mathbf{I})$ to a target data distribution $p_1 (\mathbf{x}) \coloneqq p_{\textrm{data}}(\mathbf{x})$. 
This is achieved by learning a time-dependent vector field $\mathbf{v}_\theta(\mathbf{x}_t, t)$ (parametrized with $\theta$) of a \emph{Probability Flow Ordinary Differential Equation} (PF-ODE) from $p_0$ to $p_1$. In many cases, the model $\mathbf{v}_\theta$ is conditioned with $\mathbf{c}$ such as text prompts or reference images.

The flow matching model $\mathbf{v}_\theta$ is typically trained by using a variant of simple denoising objectives. One of the representative objectives is
the conditional Flow Matching loss~\citep{flowmatching}, 
which is given as follows:
\begin{align}
\mathcal{L}_{\textrm{FM}}(\theta) = \mathbb{E}_{\mathbf{x}_t, \mathbf{c}} \big[\left\| \mathbf{v}_\theta(\mathbf{x}_t, t, \mathbf{c}) - (\mathbf{x}_1 - \mathbf{x}_0) \right\|_2^2\big]\text{,} \label{eq:fm_loss} \tag{1}
\end{align}
where $\mathbf{x}_t = (1-t) \mathbf{x}_0 + t\mathbf{x}_1$ is an interpolated sample between a Gaussian noise $\mathbf{x}_0 \sim p_0 (\mathbf{x}_0)$ and a data $\mathbf{x}_1 \sim p_1 (\mathbf{x}_1)$, and $\mathbf{c}$ is a corresponding condition of a data $\mathbf{x}_1$.
After training, we sample $\mathbf{x}_1$ by solving PF ODE from a random Gaussian noise $\mathbf{x}_0\sim p_0(\mathbf{x}_0)$ using $\mathbf{v}_\theta$.

\vspace{0.02in}\noindent
{\bf Classifier-free guidance.}~
To modulate the influence of the condition $\mathbf{c}$, classifier-free guidance (CFG)~\citep{cfg} is a well-known practice that effectively improves the generation quality and fidelity.
Specifically, by jointly training the unconditional and conditional velocity prediction models, CFG uses interpolated prediction outputs during inference:
\begin{equation*}
\mathbf{v}_\textrm{CFG}(\mathbf{x}_t, t) = \mathbf{v}_\theta(\mathbf{x}_t, t, \varnothing) + w \big(\mathbf{v}_\theta(\mathbf{x}_t, t, \mathbf{c}) - \mathbf{v}_\theta(\mathbf{x}_t, t, \varnothing)\big)\text{,} \label{eq:cfg_preliminary}
\end{equation*}
where $w \geq 1$ is the guidance scale and $\varnothing$ denotes an unconditional prediction. 
Then the CFG-modulated velocity field is used to solve PF ODE $\mathrm{d}\mathbf{x}_t = 
\mathbf{v}_\textrm{CFG}(\mathbf{x}_t, {t})\mathrm{d}t$.

\vspace{0.02in}\noindent
{\bf Latent flow matching for video generation.}
Most video generation models~\citep{hunyuan, cogvideox, wan, jin2025pyramidalflowmatchingefficient, peng2025opensora20trainingcommerciallevel, deng2025autoregressivevideogenerationvector, yu2024efficient} learn a video distribution in latent space using a two-stage framework. In the first stage, each video is spatio-temporally compressed with a 3D variational autoencoder (VAE)~\citep{kingma2013auto}, composed of an encoder $E(\cdot)$ and a decoder $G(\cdot)$. Namely, each video pixels $\mathbf{w}$ is compressed as a low-dimensional latent vector $\mathbf{x} = E(\mathbf{w})$. After that, in the second stage, a flow matching model is trained in latent space by typically leveraging the recent diffusion transformer architecture \citep{dit}. The model is trained with a full-sequence denoising objective as introduced in Eq.~\eqref{eq:fm_loss}, \emph{i.e.}, the entire video sequence is denoised at once, rather than frame by frame.
Therefore, those models lack precise control over their initial visual states, \emph{e.g.}, image-to-video (I2V) generation with an initial frame. 

To address this, I2V models~\citep{svd, xing2024dynamicrafter, wang2023videocomposer, zhang2025packing} are obtained by fine-tuning T2V models to account for the initial frame as an additional condition.
Specifically, let $\mathbf{x}_{\textrm{init}}$ be a conditioning image that is also preprocessed with VAE. Then, the goal of I2V generation is to generate a video latent vector $\mathbf{x}_1$ such that it is a natural continuation of latent image $\mathbf{x}_{\textrm{init}}$.
To condition with the initial image, several I2V models have utilized various techniques to improve image conditioning:
(1) concatenate the inflated initial frame alongside the channel dimension \citep{svd, xing2024dynamicrafter, cogvideox} and process the concatenated video, (2) add features of semantic visual encoder (\emph{e.g.}, CLIP~\citep{clip}) to the condition $\mathbf{c}$~\citep{svd, xing2024dynamicrafter, wan}, or (3) in-context conditioning with noisy initial frame~\citep{hunyuan}.
Formally, we write I2V velocity prediction model as $\mathbf{v}_\theta(\mathbf{x}_t, \mathbf{x}_{\textrm{init}}, t, \mathbf{c})$, where $\mathbf{x}_t$ is a video to predict, $\mathbf{x}_{\textrm{init}}$ is a condition image, and $\mathbf{c}$ is a text prompt.
Note that the corresponding classifier-free guidance for I2V model is given as follows: 
\vspace{-0.03in}
\begin{multline}
\mathbf{v}_\textrm{CFG-I2V}(\mathbf{x}_t, t)
  = \mathbf{v}_\theta(\mathbf{x}_t, \mathbf{x}_{\textrm{init}}, t, \varnothing)\\
  + w \Big(\mathbf{v}_\theta(\mathbf{x}_t, \mathbf{x}_{\textrm{init}}, t, \mathbf{c})
    - \mathbf{v}_\theta(\mathbf{x}_t, \mathbf{x}_{\textrm{init}}, t, \varnothing)\Big)\text{.}
\tag{2}\label{eq:cfg_i2v}
\end{multline}
\vspace{-0.05in}

\vspace{0.02in}\noindent
{\bf Suppressed motion dynamics in I2V generation.}~
Recently, several works have identified the static bias in I2V models that arises from over-conditioning and proposed various remedies~\citep{identifying, tian2025extrapolating, song2025history, flashi2v}.
\citet{identifying} trains motion-specific module to enhance motion of I2V models, and introduces an inference method that initializes noise with prior distribution and begin denoising at an earlier timestep.
Subsequently, \citet{tian2025extrapolating} controls motion strength via model merging and extrapolation, and \citet{flashi2v} proposed an I2V fine-tuning method via latent-shifting and Fourier guidance to improve motion dynamics.
Our work also targets image over-conditioning in I2V generation, but we focus on designing guidance technique that avoids image over-conditioning by adaptively removing high-frequency signals of input image.
Note that \citet{song2025history} proposed history guidance, employing partially noised frames for continued video generation, but this is only applicable to models trained with diffusion forcing~\citep{chen2024diffusion, huang2025selfforcing, cui2025selfforcingminutescalehighqualityvideo}, whereas our approach is applicable to any I2V model.

\begin{figure*}[t]
  \centering
  \begin{subfigure}[t]{0.35\textwidth}
    \centering
    \includegraphics[height=5.cm]{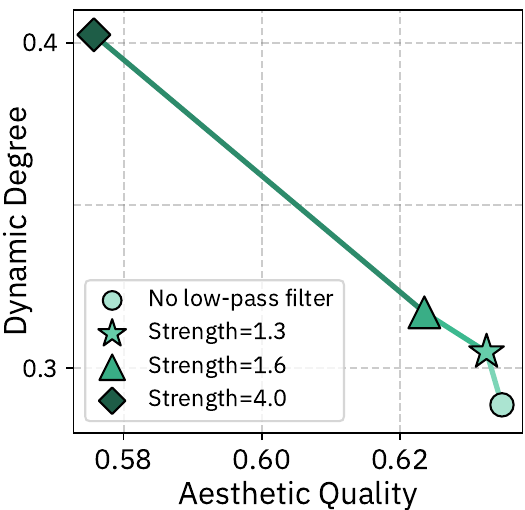}
    \caption{Video  vs. low-pass filter}
    \label{fig:t2v_to_i2v_suppression_restoration_a}
  \end{subfigure}
  \hfill
  \begin{subfigure}[t]{0.64\textwidth}
    \centering
    \includegraphics[height=5.cm]{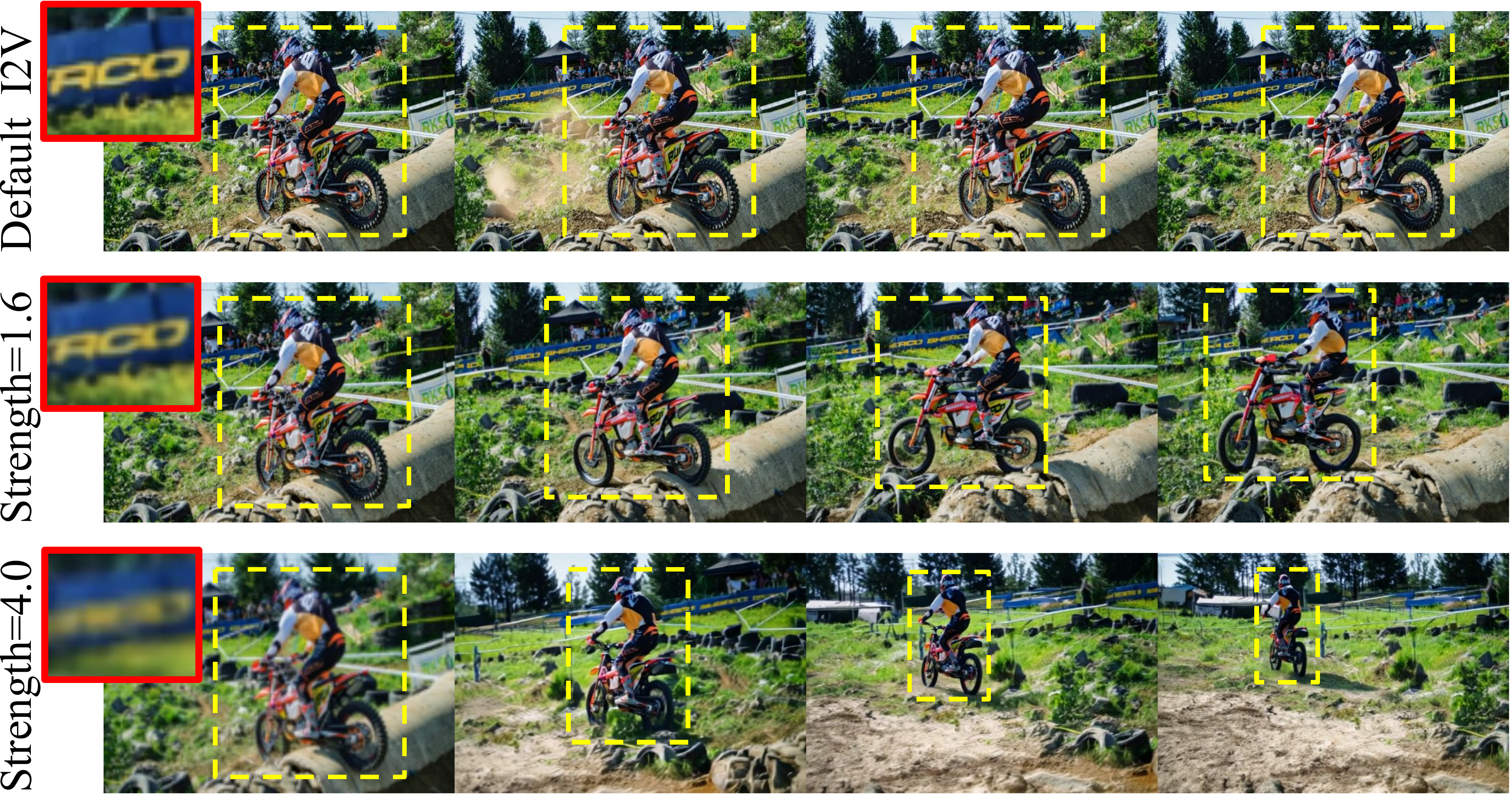}
    \caption{Visual effect of low-pass filtering on conditioning image}
    \label{fig:t2v_to_i2v_suppression_restoration_b}
  \end{subfigure}
  \caption{
    {\bf Low-pass filtering improves motion dynamics.}
    (a) We plot the dynamic degree of an I2V model (Wan 2.1~\citep{wan}) by applying low-pass filter (\emph{e.g.}, downsampling) to the input image. We observe that dynamic degree (VBench~\citep{vbench} metric which quantifies dynamicness) increases and aesthetic quality (VBench~\citep{vbench} metric which measures per-frame image quality) decreases as we use stronger low-pass filtering.
    (b) We visualize the frames when applying low-pass filtering to the input image. 
    While the videos become more dynamic using stronger low-pass filters, it sacrifices video quality as the model receives a blurry image as input (highlighted in red). 
  }
  \label{fig:t2v_to_i2v_suppression_restoration}
  \vspace{-7pt}
\end{figure*}

\section{Proposed Method}
\subsection{Suppressed motion dynamics in I2V generation} \label{sec:challenges}
In this section, we systematically investigate the root of {\em suppressed motion dynamics} in image-to-video (I2V) models—a phenomenon where the generated video exhibits minimal temporal variation, even when prompted with dynamic motion. To this end, we first identify and quantify the presence of suppressed motion, then formulate a hypothesis about its underlying cause which we verify via diagnostic experiments, and finally propose potential remedies.

\vspace{0.02in}\noindent{\bf Observation: T2V-I2V motion dynamics gap.}~
To isolate and quantify the effect of suppressed motion, we begin by comparing text-to-video (T2V) and image-to-video (I2V) models that share the same architecture and training setup.
We generate videos using T2V models, then reuse their first frames as input for I2V generation. 
This enables a fair and controlled comparison where any differences can be attributed primarily to the introduction of the conditioning image in I2V models.
We use Wan 2.1~\citep{wan}, a state-of-the-art open-source video model, which has both T2V and I2V variants.
For evaluation, we use the prompts and metrics from VBench~\citep{vbench} for dynamic degree and video quality.


\begin{table}[h!]
    \centering\small
    \setlength\tabcolsep{1.9pt}
    \begin{tabular}{lccccccc}
    \toprule
    Model & \makecell{Dynamic\\Degree} & \makecell{Subj.\\Cons.} &
    \makecell{Aesthetic\\Quality} & \makecell{Imaging\\Quality} &
    \makecell{Motion\\Smooth.} & \makecell{Temporal\\Flicker} \\
    \midrule
    T2V & 39.4 & 97.1 & 59.5 & 68.3 & 99.0 & 98.4 \\
    I2V & 32.1 & 95.2 & 56.8 & 68.4 & 99.3 & 98.7 \\
    \midrule
    $\Delta$ & {\bf -18.6\%} & -2.0\% & -4.6\% & +0.1\% & +0.3\% & +0.3\% \\
    \bottomrule
    \end{tabular}
    \caption{{\bf Motion suppression from T2V to I2V models}. Using VBench~\citep{vbench} prompts, we generate videos with Wan 2.1~\citep{wan} T2V model, and their first frames were used as I2V input to generate videos. Results show I2V models have notably less motion (dynamic degree), while other quality-related metrics remain similar.}
    \label{tab:t2v_to_i2v_suppression}
\end{table}
Tab.~\ref{tab:t2v_to_i2v_suppression} compares VBench metrics between the T2V and I2V model.
Note the 18.6$\%$ drop in dynamic degree of I2V model compared to T2V, while the video quality metrics (all except dynamic degree) remain comparable.
This result shows that I2V models fall short in generating dynamic videos compared to T2V ones, which might be introduced due to the conditioning mechanism of I2V models that is inserted during adaptation from T2V models.
Detailed explanation of each VBench metric can be found in Appendix~\ref{app:experimental_details}.

\vspace{0.05in}\noindent{\bf Hypothesis: exposure to high-frequency signals.}~
We hypothesize that the suppression of motion arises from the over-exposure to {\em high-frequency elements} during early generation stages, which disrupts the coarse-to-fine nature of the generative process.
Specifically, we observe that I2V models suffer from ``shortcut'' during generation, where fine-grained details of input image lock in the early generation stages, confining the generation trajectory from the beginning (see Fig.~\ref{fig:shortcut} top row).
As such, the loss of flexibility in the generation trajectory hinders the formation of temporal variations,
resulting in a static video.

\vspace{0.1in}\noindent
{\bf Diagnosis: low-pass filtering alleviates exposure.}~
To handle suppressed motion dynamics, we claim that applying low-pass filter to the condition image relieves the over-conditioning by removing the high-frequency features. 
To show this, we perform a simple diagnostic experiment that applies a low-pass filter (\emph{e.g.}, downsampling) to the input image and generate videos with an I2V model by varying strength. 
We use VBench test set and compute the average dynamic degree for the generated videos using each low-pass filter strength. 
Fig.~\ref{fig:t2v_to_i2v_suppression_restoration} shows the results.
We observe that applying low-pass filter to the input image (\emph{i.e.}, removing high-frequency components) improves the dynamicness of generated videos, which increases monotonically with respect to the strength.
Note that this supports our hypothesis that high-frequency details in the conditioning image hinder synthesizing dynamic motion. 
However, as shown in Fig.~\ref{fig:t2v_to_i2v_suppression_restoration}, na\"ive application of low-pass filter introduces a trade-off, as it sacrifices fidelity, making it impossible to perfectly reconstruct the original image. 
We report additional results using different I2V models in Appendix~\ref{app:additional_results_suppression}.

\vspace{0.02in}\noindent{\bf Visualizing and mitigating the shortcut effect.}~ 
To corroborate the existence of shortcut effect, we inspect the internal representations of I2V models. Specifically, we extract feature maps from the intermediate layer of an I2V model (Wan 2.1~\citep{wan}), and visualize them by using principal component analysis (PCA) to convert into RGB images, similar to DINOv2~\citep{dinov2} and REPA~\citep{repa}. As shown in Fig.~\ref{fig:shortcut} (first row), the feature map locks onto fine details of the input image after just one denoising step (at $t$=0.02 out of 50 steps), limiting the flexibility of subsequent generation steps and resulting in static video. In contrast, applying a low-pass filter to the input image (Fig.~\ref{fig:shortcut} second row) prevents this shortcut, and allows a more varied trajectory that results in a more dynamic video. The early occurrence of this shortcut effect, together with the fact that low-pass filtering mitigates this effect, motivates us to apply filtering only at the beginning stages of generation. We visualize feature maps of other I2V generation results and report the results similarly to Fig.~\ref{fig:shortcut} in Appendix~\ref{app:additional_results_fmap}.

In summary, our findings suggest that the suppressed motion in I2V models (compared to T2V models) results mainly from high-frequency details in the conditioning image guiding generation prematurely into a static ``shortcut,'' especially in the crucial early stages when motion patterns should emerge. While low-pass filtering effectively prevents this shortcut, it comes at the cost of reduced image fidelity. This trade-off motivates us to develop a method, introduced in the next section, that restores T2V-level motion dynamics while preserving video quality and image fidelity.

\vspace{0.1in}
\begin{figure*}[t]
  \centering
  \includegraphics[width=.98\linewidth]{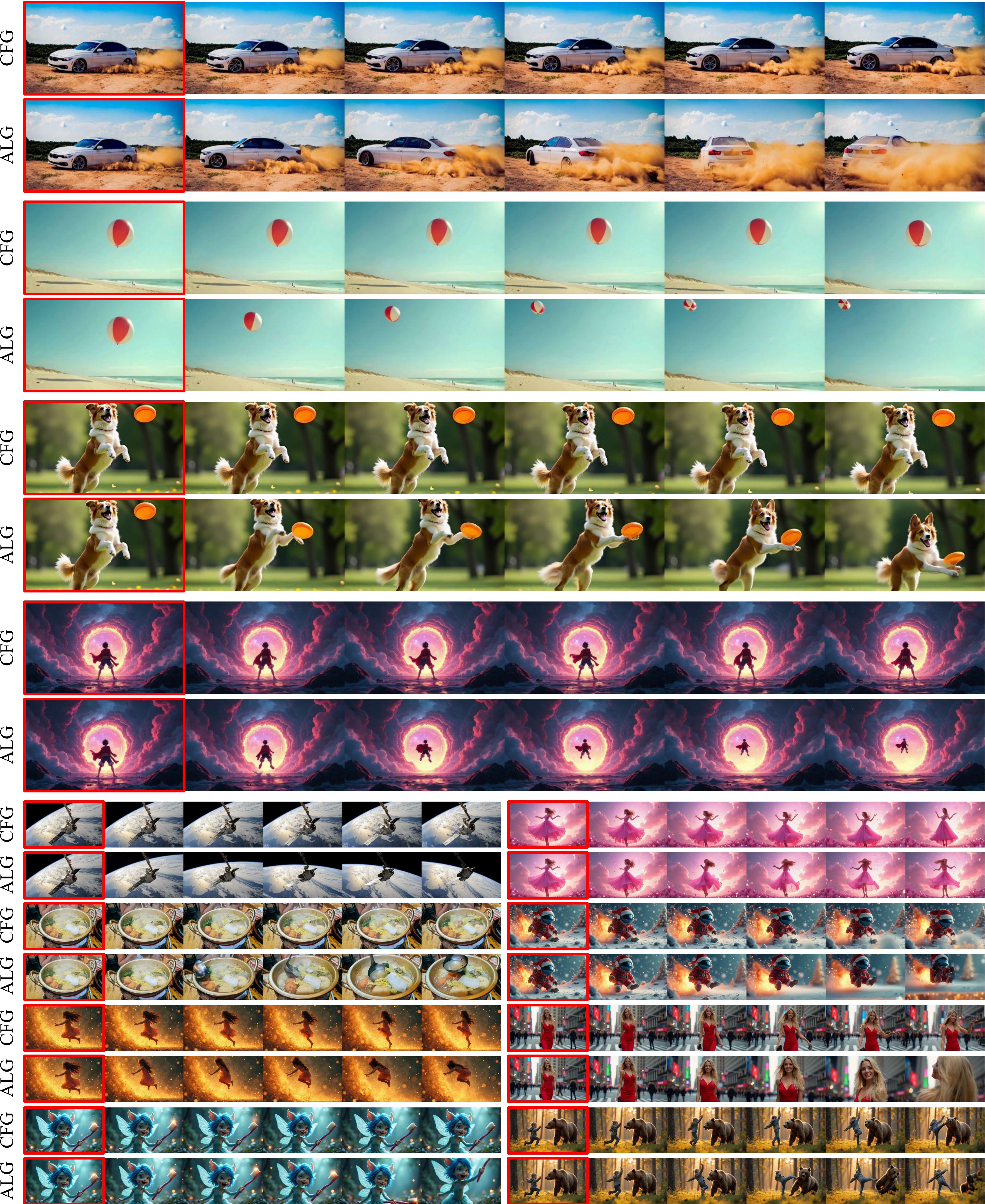}
  \vspace{-5pt}
  \caption{
  {\bf Qualitative comparison between \method and CFG.}~ 
  We provide visual comparison between the videos generated by using default image-to-video generation method (CFG) and our method (\method).
  The input conditioning frames are denoted with red outline.
  We observe that the videos using \method show more dynamic motion (\emph{e.g.}, larger object movement, animal movement, or human action, and more complex background movements).
  The list of prompts and models used for each video is included in the supplementary material.
  }
  \label{fig:main_qual}
  \vspace{-15pt}
\end{figure*}

\subsection{Adaptive Low-Pass Guidance} \label{sec:method}
Based on our analysis in Sec.~\ref{sec:challenges}, we present a method that alleviates the suppressed motion dynamics of I2V generation. Our main goal is to enhance the motion of I2V models comparable to T2V models, while preserving fidelity to the input image. To this end, we propose \emph{adaptive low-pass guidance} (\method), a simple training-free inference technique for I2V models that controls the amount of high-frequency components in the input image during sampling.

Specifically, \method applies the low-pass filter to the image, but dynamically modulates its strength with respect to the timestep $t$. 
To use only low-frequency information at the early stage, we apply stronger low-pass filtering at the earlier phase (\emph{i.e.}, $t\approx 0$) and progressively reduce the strength, which leads to using the original high-frequency image condition at the latter phase (\emph{i.e.}, $t\approx 1$). 
Thus, by exposing the original image at the later stage, the model can reconstruct the high-frequency details of the image from intermediate states that are dynamic but lack fine-grained details.

\begin{table*}[t!]
  \centering
  \begin{minipage}[t]{0.485\textwidth}
    \centering\footnotesize
    \setlength\tabcolsep{3.3pt}
    \begin{tabular}{lc>{\columncolor[HTML]{DAE7ED}}ccccccc}
    \toprule
    {Model} & {Method} & \makecell{Dynamic\\Degree} & \makecell{\\Avg.} & \makecell{VBench\\QS} & \makecell{\\I2V} & \makecell{DOVER} &\makecell{Vision\\Reward} \\
    \midrule
    \multirow{2}{*}{Wan 2.2}
    & CFG & 31.7 & 79.6 & \textbf{85.4} & 98.5 & 0.635 & \textbf{0.183} \\
    & \method & \textbf{39.0} & \textbf{80.5} & 85.2 & \textbf{98.5} & \textbf{0.637} & 0.182 \\
    \midrule
    \multirow{2}{*}{Wan 2.1}
    & CFG & 28.9 & 79.1 & \textbf{85.3} & \textbf{98.3} & \textbf{0.618} & \textbf{0.179} \\
    & \method & \textbf{39.4} & \textbf{80.0} & 84.5 & 98.0 & 0.614 & 0.176 \\
    \midrule
    \multirow{2}{*}{\makecell{LTX-\\Video}}
    & CFG & 15.5 & 77.8 & \textbf{85.9} & \textbf{99.1} & 0.625 & 0.175 \\
    & \method & \textbf{21.5} & \textbf{78.2} & 85.4 & 98.9 & \textbf{0.626} & \textbf{0.175} \\
    \midrule
\end{tabular}

    \captionof{table}{
    {\bf Comparison across various I2V models}. 
    Using Wan 2.1/2.2~\citep{wan}, and LTX-Video on VBench~\citep{vbench}, videos from \method show higher dynamicness (\textcolor[HTML]{1f6e94}{blue}) compared to baseline (CFG), while quality metrics (all except Dynamic Degree) remain similar.
    }
    \label{tab:main_results}
  \end{minipage}\hfill
  \begin{minipage}[t]{0.485\textwidth}
    \centering\footnotesize
    \setlength\tabcolsep{2.7pt}
    \begin{tabular}{lc>{\columncolor[HTML]{DAE7ED}}cccccc}
    \toprule
    \makecell{Benchmark\\Dataset} & {Method} & \makecell{Dynamic\\Degree} & \makecell{\\Avg.} & \makecell{VBench\\QS} & \makecell{\\I2V} & \makecell{DOVER} & \makecell{Vision\\Reward} \\
    \midrule
    \multirow{2}{*}{VBench}
    & CFG & 31.7 & 79.6 & \textbf{85.4} & 98.5 & 0.635 & \textbf{0.183} \\
    & \method & \textbf{39.0} & \textbf{80.5} & 85.2 & \textbf{98.5} & \textbf{0.637} & 0.182 \\
    \midrule
    \multirow{2}{*}{PVD}
    & CFG & 65.0 & 79.4 & 79.6 & 94.2 & 0.484 & 0.145 \\
    & \method & \textbf{69.0} & \textbf{80.3} & \textbf{79.6} & \textbf{95.0} & \textbf{0.512} & \textbf{0.145} \\
    \midrule
    \multirow{2}{*}{VidProM}
    & CFG & 27.3 & 79.1 & 85.6 & \textbf{98.2} & \textbf{0.658} & 0.096 \\
    & \method & \textbf{30.5} & \textbf{79.5} & \textbf{85.6} & 98.0 & 0.657 & \textbf{0.096} \\
    \midrule
\end{tabular}

    \captionof{table}{
    {\bf Comparison across various benchmarks}. 
    Across three datasets (VBench~\citep{vbench}, PE Video Dataset (PVD)~\citep{pvd}, and VidProM~\citep{vidprom}), and using Wan 2.2~\citep{wan}, \method improves motion dynamics (\textcolor[HTML]{1f6e94}{blue}) over the baseline (CFG) while maintaining quality.
    }
    \label{tab:acrossdatasets}
  \end{minipage}
  \vspace{-2mm}
\end{table*}
\vspace{0.02in}

\vspace{0.1in}
\noindent
{\bf Guidance.}~
Formally, let $\mathbf{x}_{\textrm{init}}^{(t)} = \mathcal{F}_{\textrm{LP}}\big(\mathbf{x}_\textrm{init}, \kappa(t)\big)$ be a low-pass filtered version of the original latent image $\mathbf{x}_{\textrm{init}}$ using pre-defined low-pass filter $\mathcal{F}_{\textrm{LP}}$ (\emph{e.g.}, Gaussian blur or bilinear resizing), where a strength factor $\kappa:[0,1]\rightarrow \mathbb{R}$ is given as a decreasing function of timestep $t$. Note that $\mathcal{F}_{\textrm{LP}}\big(\mathbf{x}_\textrm{init}, 0\big) = \mathbf{x}_{\textrm{init}}$ since the filter strength is zero.
Thus, one can prevent the sampling trajectory becoming ``shortcut solution'' by exposing filtered latent, and also allow the model to reconstruct fine details by exposing the original unfiltered latent in the later timesteps.
We use this adaptive condition $\mathbf{x}_{\textrm{init}}^{(t)}$ in the CFG formula (Eq.~\eqref{eq:cfg_i2v}), namely:
\vspace{-0.10in}
\begin{multline}
\mathbf{v}_{\textrm{\method}}(\mathbf{x}_t, t)
  = \mathbf{v}_\theta(\mathbf{x}_t, \mathbf{x}_{\textrm{init}}, t, \varnothing)\\
  + w \big(
      \mathbf{v}_\theta(\mathbf{x}_t, \mathbf{x}_{\textrm{init}}^{(t)}, t, \mathbf{c})
      - \mathbf{v}_\theta(\mathbf{x}_t, \mathbf{x}_{\textrm{init}}^{(t)}, t, \varnothing)
    \big)\text{.}
\tag{3}\label{eq:cfg_lpg}
\end{multline}

One important design choice of the formulation of our method in Eq.~\eqref{eq:cfg_lpg} is that we use ${\mathbf{x}^{(t)}_\textrm{init}}$ only for the latter two terms, leaving the first unconditional term $\mathbf{v}_\theta(\mathbf{x}_t, \mathbf{x}_{\textrm{init}}, t, \varnothing)$ with the original input image ($\mathbf{x}_\textrm{init}$).
This choice allows us to balance enhanced motion with fidelity to the input conditioning image. 
To better understand its effect, Eq.~\eqref{eq:cfg_lpg} can be algebraically rearranged into the following equivalent form:
\begin{multline*}
   \mathbf{v}_\textrm{\method}(\mathbf{x}_t, t)\\
   =\underbrace{\left[
      \begin{aligned}
        &\mathbf{v}_\theta(\mathbf{x}_t, \mathbf{x}^{(t)}_\mathrm{init}, t, \varnothing)\\
        &+ (w-1) \Big(\mathbf{v}_\theta(\mathbf{x}_t, \mathbf{x}^{(t)}_\mathrm{init}, t, \mathbf{c})
        - \mathbf{v}_\theta(\mathbf{x}_t, \mathbf{x}^{(t)}_\mathrm{init}, t, \varnothing)\Big)
      \end{aligned}
    \right]}_\text{(a) Motion-Enhanced CFG} \\
  + \underbrace{\Big(\mathbf{v}_\theta(\mathbf{x}_t, \mathbf{x}_\mathrm{init}, t, \varnothing)
    - \mathbf{v}_\theta(\mathbf{x}_t, \mathbf{x}^{(t)}_\mathrm{init}, t, \varnothing)\Big)}_\text{(b) Fidelity Correction}.
\end{multline*}
Term (a) is the standard CFG sampling, where both unconditional and conditional predictions use the low-pass filtered image $\mathbf{x}_\textrm{init}^{(t)}$, promoting dynamic motion.
Term (b) guides the sampling towards the high-frequency visual information that might be lost if only term (a) were used.

Empirically, in our experiments, using the low-pass filtered image in all three terms indeed resulted in less stable generation results, often characterized by distorted visuals, reduced spatial coherence, or sudden scene transitions. Visual examples can be found in Appendix~\ref{app:design_choice}.

\vspace{0.1in}
\noindent
{\bf Choice of $\kappa(t)$.}~
In our formulation, $\kappa(t)$ controls the low-pass filter strength at timestep $t$.
As the main purpose of our formulation in Eq.~\eqref{eq:cfg_lpg} is to apply a stronger low-pass filter in earlier steps (\emph{i.e.}, $t \approx 0$) and reduce strength later on, any choice of $\kappa(t)$ that satisfies this condition suffices.

A simple exemplary choice is a step function which maps to a large initial filter strength $\kappa_\ast>0$ when $t \approx 0$ and drops to $0$ later on.
Such step function can be defined as
\begin{align}
\kappa(t) = \begin{cases}
\kappa_{\ast} & \text{if } t < t_\textrm{trans} \\
0 & \text{if } t \ge t_\textrm{trans},
\end{cases} \tag{4} \label{eq:step_schedule_fm}
\end{align}
with a transition point hyperparameter $t_\textrm{trans}\in(0,1)$ and initial filter strength hyperparameter $\kappa_{\ast}>0$.

In practice, we find that any choice of $\kappa(t)$ for \method can improve the motion dynamics of the generated videos, as long as it assigns high filter strength at the early sampling steps, exposing the initial sampling process to the low-pass filtered input image latent.\footnote{We find denoising for 1-2 steps with clean latents before switching to the low-pass filtered latent, thereby slightly delaying the exposure to the filtered latent, to improve quality, at a slight cost to dynamicness.}
On the other hand, a prolonged exposure to the low-pass filtered input image latent (\emph{e.g.}, large $t_\textrm{trans}$ in the step function) often led to a loss of input image fidelity.
See Sec.~\ref{sec:experiments} for more implementation details of our method, and Sec.~\ref{sec:component_analysis} for a detailed analysis of how each of the hyperparameters (\emph{e.g.}, $\kappa$ and $t_\textrm{trans}$) affect video motion and quality.
\section{Main Experiments}\label{sec:experiments}
\vspace{0.08in}
{\bf Models.}~
We apply adaptive low-pass guidance (\method) to recent state-of-the-art open-source I2V models, mainly Wan 2.2~\citep{wan}, and additionally Wan 2.1~\citep{wan} and LTX-Video~\citep{ltx}.
The primary baseline of our experiment is the standard generation using the official checkpoints without \method.
Details on models can be found in Appendix~\ref{app:experimental_details}.

\vspace{0.02in}\noindent
{\bf Evaluation datasets.}~
For comprehensive evaluation, we curate three datasets: VBench test set~\citep{vbench}, PE Video Dataset (PVD)~\citep{pvd}, and VidProM~\citep{vidprom}.
VBench test set consists of image-caption pairs, which we use all except those for assessing camera motion and background quality.
From PVD, a video-caption dataset, we randomly sample 100 videos and use the first frames to build an image-caption pair dataset.
Finally, from VidProM, a text-to-video prompt dataset, we randomly sample 750 prompts and generate corresponding images using FLUX.1-dev~\citep{flux}.
More details about datasets can be found in Appendix~\ref{app:evaluation_set}.

\vspace{0.02in}\noindent
{\bf Evaluation Metrics.}~
To measure the dynamicness and quality of videos, we employ various metric suites.
Our main goal is to evaluate \method's ability to mitigate motion suppression, measured by {\em Dynamic Degree}, a VBench~\citep{vbench} metric.
To monitor input image fidelity and visual quality, we additionally report quality-related metrics.
VBench-QS (Quality Score) evaluates key visual aspects of a video by averaging several VBench metrics, including Subject Consistency, Motion Smoothness, Aesthetic Quality, Imaging Quality, and Temporal Flicker.\footnote{We omit metrics assessing background quality and camera motion due to the high cost of evaluation.}
Also, VBench-I2V (I2V Subject Consistency), measures input image fidelity via DINO~\citep{dino} similarity.
Finally, VBench-Avg. averages all metrics (including Dynamic Degree) to assess I2V performance.
We further include two additional metrics: VisionReward~\citep{visionreward}, a VLM-based human-aligned video reward model, and DOVER~\citep{DOVER}, a video quality evaluator for aesthetic and technical quality (trained using human judgment data).
Details on evaluation can be found in Appendix~\ref{app:evaluation_set}.

\vspace{0.02in}\noindent
{\bf \method implementation details.}~
For the low-pass filter $\mathcal{F}_\textrm{LP}$, we apply a bilinear downsampling to the input image latent $\mathbf{x}_\textrm{init}$ followed by a bilinear upsampling back to the original resolution.
Initial filter strength $\kappa_*$ is set to the downsampling factor of 2.5 by default, but is adjusted for some models with higher or lower resolution.
By default, we set the transition time $t_\textrm{trans}$=0.1 for the step function schedule (\emph{i.e.}, Eq.~\eqref{eq:step_schedule_fm}), meaning that low-pass filtering is applied to first $10\%$ of the denoising timesteps.
Finally, as mentioned in Sec.~\ref{sec:method}, we additionally denoise using clean latent for the first 2 steps, and also override the first-frame latent with the clean version at decode-time. We find both techniques to improve video quality without introducing any additional overhead.
The full implementation details and hyperparameters of our experiments can be found in Appendix~\ref{app:method_details}.

\vspace{0.02in}\noindent
{\bf Qualitative results.}~
Fig.~\ref{fig:main_qual} presents a qualitative comparison between the default I2V generation method (CFG) and our method (\method).
Our method produces videos with more dynamic motion, whereas the default CFG method tends to generate more static videos, consistent with the results shown in Tab.~\ref{tab:main_results} and Tab.~\ref{tab:acrossdatasets}.
For example, objects or vehicles show more dynamic movement, human and animal actions appear more active, and scene transitions are more complex.
Details about video generation, including prompts used for video generation can be found in Appendix~\ref{app:experimental_details}. 
Additional qualitative examples can be found in Appendix~\ref{app:additional_results_qualitative}.

\vspace{0.02in}\noindent
{\bf Quantitative results.}~
Tab.~\ref{tab:main_results} reports the 
evaluation results comparing the I2V performance using default generation baseline (\emph{i.e.}, CFG) and our method (\method) using three I2V models (Wan 2.2, Wan 2.1, LTX-Video) on VBench. 
We observe that \method shows consistent improvement on video dynamicness (via Dynamic Degree), crucially with comparable or even often more favorable quality metrics (VBench-QS, VBench-I2V, DOVER, VisionReward).
As a result, VBench-Avg. improves in \method compared to baseline across all I2V models and benchmark datasets.
We observe similar results consistent across various benchmark datasets (VBench, PVD, VidProM), as shown in Tab.~\ref{tab:acrossdatasets}.

\begin{figure*}[t]
  \begin{subfigure}[t]{0.32\textwidth}
    \centering
    \includegraphics[height=4.5cm]{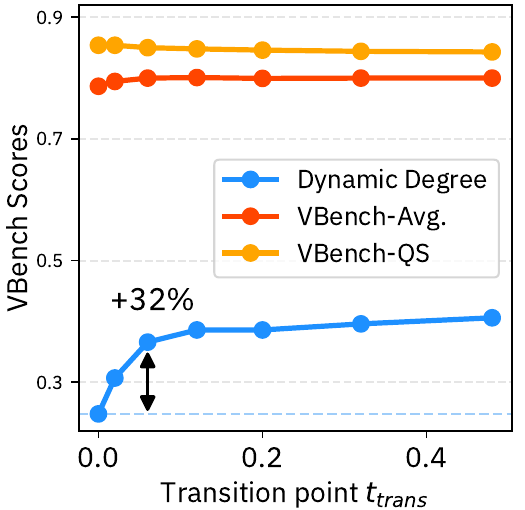}
    \caption{Transition point $t_\textrm{trans}$ of \method}
    \label{fig:component_trans}
  \end{subfigure}
  \hfill
  \begin{subfigure}[t]{0.32\textwidth}
    \centering
    \includegraphics[height=4.5cm]{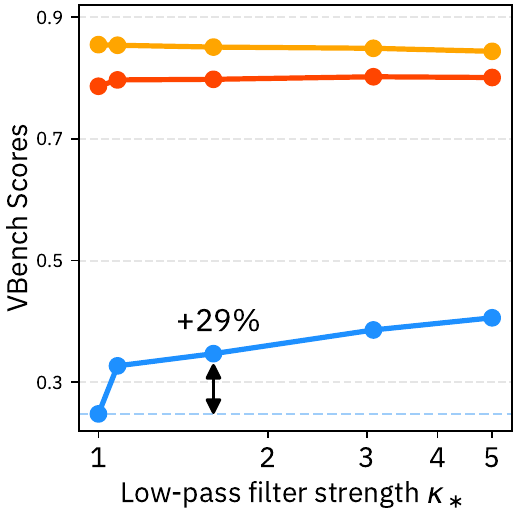}
    \caption{Filter strength $\kappa_\ast$ of \method}
    \label{fig:component_strength}
  \end{subfigure}
  \hfill
  \begin{subfigure}[t]{0.32\textwidth}
    \centering
    \includegraphics[height=4.5cm]{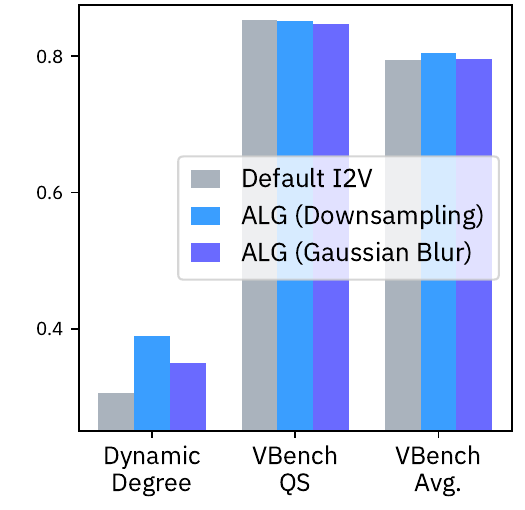}
    \caption{Low-pass filter type of \method}
    \label{fig:component_filter_type}
  \end{subfigure}
  \caption{{\bf Component analysis with VBench-I2V.} (a) As $t_\textrm{trans}$ increases from 0, dynamic degree increases rapidly, while quality metrics remain stable or slightly drops. This indicates that high-frequency signals prevent dynamic motions from forming in early generation steps. (b) Increasing the initial low-pass filter strength $\kappa_\ast$ shows that \method can enhance dynamicness without significantly sacrificing video quality. (c) Both bilinear downsampling and Gaussian blur show enhanced dynamics over default I2V.}
  \label{fig:component}
\end{figure*}
\section{Analysis}\label{sec:component_analysis}
\begin{table}[t]
    \centering\footnotesize
    \vspace{2pt}
    \setlength\tabcolsep{4pt}
    \begin{tabular}{l>{\columncolor[HTML]{DAE7ED}}ccccccc}
        \toprule
        {Method} & \makecell{Dynamic\\Degree} & \makecell{\\Avg.} & \makecell{VBench\\QS} & \makecell{\\I2V} & \makecell{DOVER} & \makecell{Vision\\Reward} \\
        \midrule
        CFG. & 31.7 & 79.6 & \textbf{85.4} & 98.5  & 0.635 & \textbf{0.183} \\
        \method. & \textbf{39.0} & \textbf{80.5} & 85.2 & \textbf{98.5} & \textbf{0.637} & 0.182 \\
        \midrule
        CFG + \textit{Aug.} & 38.6 & 80.2 & \textbf{85.0} & 98.2 & 0.635 & 0.168 \\
        \method + \textit{Aug.} & \textbf{43.5} & \textbf{80.8} & 84.7 & \textbf{98.2} & \textbf{0.636} & \textbf{0.171} \\
        \midrule
    \end{tabular}
    \caption{{\bf Comparison with motion-augmented prompts on VBench-I2V~\citep{vbench} with Wan~2.2}.
    Prompts augmented via Gemini~2.5~\citep{gemini} (\textit{Aug.}) help both methods. Notably, \method surpasses CFG baseline on Dynamic Degree even without augmentation.
    }
    \label{tab:tab_enhanced}
\end{table}

\vspace{0.02in}
{\bf Motion-augmented prompts.}~
Orthogonally to \method, image-to-video input text prompts can be altered to become more dynamic by emphasizing motion in them. We use Gemini 2.5~\citep{gemini} to motion-enhance VBench test set prompts and compare our method (\method) and baseline (CFG). Tab.~\ref{tab:tab_enhanced} shows that both methods benefit from the augmentation. It is worth noting that \method, even without augmentation, is more dynamic compared to CFG.
More details about prompt augmentation can be found in  Appendix~\ref{app:experimental_details}.

\vspace{0.02in}\noindent
{\bf Transition time }$t_{\textrm{trans}}${\bf .}~
To demonstrate the effect of $t_{\textrm{trans}}$, we compute Dynamic Degree, VBench-QS (average of quality-related VBench metrics), and VBench-Avg. (average of all metrics, including Dynamic Degree) following Sec.~\ref{sec:experiments} by varying the transition time.
The results are shown in Fig.~\ref{fig:component_trans}.
We observe that Dynamic Degree increases 32\% when $t_{\textrm{trans}}$ = 0.06, showing that low-pass filtering at the early step is enough to enhance motion, which aligns with our findings in Sec.~\ref{sec:challenges}.
Moreover, we see that other metrics such as VBench-QS and VBench-Avg. remain largely stable without any quick decrease, allowing us to enhance motion dynamics without a significant loss of quality.

\vspace{0.02in}\noindent
{\bf Initial filter strength }$\kappa_*${\bf .}~
Next, we demonstrate the effect of various initial strengths for low-pass filtering (bilinear downsampling).
As presented in Fig.~\ref{fig:component_strength}, we find that increasing the filter strength generally makes the video more dynamic with a diminishing gain of dynamicness. 
We find the increased video motion comes with a minimal cost of video quality. For instance, at $\kappa_\ast=1.6$, Dynamic Degree increases by 29\% while VBench-QS decreases by 0.5\%. 
As a result, the overall score (VBench-Avg.) increases.

\vspace{0.02in}\noindent
{\bf Low-pass filter type.}~
We additionally evaluate \method using Gaussian blur to test a different type of low-pass filter than bilinear downsampling.
As shown in Fig.~\ref{fig:component_filter_type}, applying Gaussian blur shows similar trends to downsampling that improves dynamicness, while other metrics slightly drop.
We note that the gain is smaller than downsampling, which we hypothesize that this is because downsampling followed by upsampling removes fine details more aggressively than Gaussian blur, mainly due to the decimation of signals.

\section{Conclusion} \label{sec:conclusion}
In this work, we investigate the suppressed motion dynamics prevalent in current image-to-video (I2V) generation models.
We identify that high-frequency components within input images cause a ``shortcut'' effect, where generation trajectory prematurely locks onto the image's appearance during denoising.
We demonstrate that low-pass filtering can mitigate this issue and enhance video motion, at the cost of input image fidelity.
Building on these insights, we propose \method, which adaptively applies low-pass filtering to the reference image only during the early denoising stages to encourage motion, then reverts to the original image in later stages to ensure fidelity and quality.
Extensive evaluations demonstrated that \method significantly boosts motion dynamicness (\emph{e.g.}, average 33\% improvement on VBench across various models), while maintaining video quality.

{
    \small
    \section*{Acknowledgments} 
    This work was supported by the Institute of Information \& communications Technology Planning \& Evaluation (IITP) grant funded by the Korea government (MSIT) (RS-2019-II190075, Artificial Intelligence Graduate School Program (KAIST)), (No. RS-2024-00509279, Global AI Frontier Lab), and (RS-2025-02304967, AI Star Fellowship (KAIST)).
}
{
    \small
    \bibliographystyle{ieeenat_fullname}
    \bibliography{main}
}

{
\clearpage
\appendix
\setcounter{page}{1}
\onecolumn
\maketitlesupplementary
\label{appendix}


\section{Details about \method}\label{app:method_details}

\subsection{Implementation details of \method}

\vspace{0.05in}
\noindent
{\bf General algorithm for \method.}~
Algorithm~\ref{alg:alg} shows the general algorithm for \method applying to all models.
Note that while $\kappa(t)$ is written in the most general form possible (\emph{i.e.}, any $\kappa:[0,1]\to\mathbb{R}$), we use step function in our experiments (Sec.~\ref{sec:method}).
When $\kappa(t) = 0$, no filter is applied (\emph{i.e.}, $\mathbf{x}^{(t)}_\textrm{init} = \mathbf{x}_\textrm{init}$) and the sampling becomes equivalent to classifier-free guidance (CFG).

\begin{algorithm}[h]
\caption{Image-to-video sampling with Adaptive Low-Pass Guidance (\method)}
\label{alg:alg}
\begin{algorithmic}[1]
  \Require Denoiser \(\mathbf{v}_\theta\), encoder \(E\), decoder \(G\), input conditioning image \(\mathbf{w}_{\mathrm{init}}\), prompt \(\mathbf{c}\), guidance \(w\), low-pass filter \(\mathcal{F}_{\mathrm{LP}}\), strength schedule \(\kappa:[0,1]\!\to\!\mathbb{R}\), total inference steps \(N\)
  \State \(\mathbf{x}_{\mathrm{init}} \gets E(\mathbf{w}_{\mathrm{init}})\)  \Comment{Encode the input conditioning image}
  \State \(\mathbf{x} \sim \mathcal{N}(\mathbf{0}, \mathbf{I})\) 
  \For{\(i = 1\) to \(N\)}                 
    \State \(t \gets \tfrac{i}{N}\)       
    \State \(\mathbf{x}_{\mathrm{init}}^{(t)} \gets 
           \mathcal{F}_{\mathrm{LP}}\bigl(\mathbf{x}_{\mathrm{init}},\,\kappa(t)\bigr)\)
    \State \(\displaystyle
      \mathbf{v}_{\mathrm{\method}} \gets
      \mathbf{v}_\theta(\mathbf{x},\mathbf{x}_{\mathrm{init}},t,\varnothing)
      +w\,\Bigl[
        \mathbf{v}_\theta(\mathbf{x},\mathbf{x}_{\mathrm{init}}^{(t)},t,\mathbf{c})
        -
        \mathbf{v}_\theta(\mathbf{x},\mathbf{x}_{\mathrm{init}}^{(t)},t,\varnothing)
      \Bigr]\)
    \State \(\mathbf{x} \gets \mathrm{SolverStep}(\mathbf{x},\mathbf{v}_{\mathrm{\method}},t)\)
  \EndFor
  \State \Return \(G(\mathbf{x})\)   \Comment{Decode the final latent into video}
\end{algorithmic}
\end{algorithm}

While the implementation of \method is straightforward, subtle differences per model (Wan 2.1/2.2~\citep{wan}, LTX-Video~\citep{ltx}) exist, depending on specific model architecture and implementation. 
In most cases, $\mathbf{x}$ is a 5-dimensional tensor (batch size, frame count, channel count, width, height). 
$\mathbf{x}_\textrm{init}$ is provided as input by either concatenating to $\mathbf{x}$ along channel dimension (Wan 2.1/2.2) or replacing the first input token with $\mathbf{x}_\textrm{init}$ (LTX-Video). 
Note that as $\mathbf{x}_\textrm{init}$ has a different shape (\emph{i.e.}, frame count is 1) than $\mathbf{x}$, we expand its shape via zero padding.
Implementation details of \method for each model is explained below.

\vspace{0.05in}
\noindent
{\bf Implementation of \method for Wan 2.1.}~
Wan 2.1~\citep{wan} is a flow-matching model based on a diffusion transformer (DiT; \citep{dit}) backbone, fine-tuned from its pre-trained base text-to-video model checkpoint to perform an image-to-video generation task. 
The conditioning image is first encoded into the VAE latent space using the VAE encoder, and is incorporated into the input via zero-padding followed by channel-wise concatenation.
We implement \method by simply applying low-pass filter to the input conditioning image latent before performing zero-padding.
Wan 2.1 has an additional input to the DiT---the CLIP embedding of the conditioning image. 
As the purpose of the CLIP embedding is to provide a high-level semantic information (and not fine-grained details of the image such as small edges), we do not apply low-pass filter to the input for the CLIP encoder and just use the original image.
We implement \method under the official diffusers codebase of Wan 2.1.

\vspace{0.05in}
\noindent
{\bf Implementation of \method for Wan 2.2.}~
Wan 2.2~\citep{wan} is likewise a DiT~\citep{dit}-based flow-matching image-to-video model, but with a two-stage denoiser (one for high-noise steps and the other for low-noise steps) and an explicit first-frame masking mechanism.
Similarly to Wan 2.1, the input image is first encoded into the VAE latent space, temporally zero-padded to match the target number of frames, and concatenated channel-wise.
\method is implemented similarly to Wan 2.1 as well, and is applied to the VAE-encoded input image before concatenation.
One difference is that the CLIP embedding is disabled for the 14B sized Wan 2.2 model. 
We implement \method on top of the official diffusers implementation of Wan 2.2.

\vspace{0.05in}
\noindent
{\bf Implementation of \method for LTX-Video.}~
LTX-Video~\citep{ltx} is a DiT~\citep{dit}-based flow-matching model. Distinctively from Wan 2.1 and Wan 2.2, LTX-Video incorporates the input conditioning image by substituting the first frame of the noisy video latent with the clean conditioning image latent at each denoising step.
At each denoising step, a scheduled noise is added to the conditioning image latent (which is used as the first frame). 
To integrate \method into LTX-Video, we apply \method by using low-pass filtered conditioning image latent as the first frame during the early steps (\emph{i.e.}, before $t\in [0, t_\textrm{trans})$) and switching to the original conditioning image with a scheduled noise added to it thereafter (\emph{i.e.}, $t\in [t_\textrm{trans}, 1)$).

\vspace{0.05in}
\noindent
{\bf Low-pass filter implementation.}~
In our \method experiments, we use downsampling followed by upsampling as our choice of low-pass filter.
Specifically, we first bilinearly downsample the original latent into a smaller latent size (so that the latent width becomes $\textrm{latent\_width}/\kappa(t)$), then upsample it back to the original latent size. 
While there are various possible choices of interpolation functions other than bilinear interpolation, we use it in our main experiments due to its simplicity.

\vspace{0.05in}
\noindent
{\bf Additional techniques.}~
As discussed in Sec.~\ref{sec:method}, we apply two additional techniques that we find to improve video quality.
First, after generation ends, the first frame of the final latent is overriden with the clean input latent. This is similar to \texttt{expand\_timesteps} feature enabled by default for Wan 2.2 5B model (not in 14B, which we use), which overrides the first frame of the noisy latent with the clean latent during denoising. 
Our technique is distinct in that we override after denoising is entirely over, instead of during denoising.
We empirically find this technique to slightly improve video quality, while using \texttt{expand\_timesteps} feature introduces noticeable artifacts.
Additionally, we find that denoising using the clean input latent at the beginning of the denoising process for 1 or 2 steps to help improve quality.
Note that this does not affect the duration of denoising steps using the low-pass filtered latent.
Instead, it merely \textit{delays} the exposure of the model to the low-pass filtered latent slightly.
These two techniques do not incur any additional computational overhead.
The configuration for these techniques are described in Tab.~\ref{tab:app_models_modified} (\emph{i.e.}, first-frame override, low-pass filter delaying).

\begin{table}[t]
\centering\small
\vspace{2pt}
\begin{tabular}{l c cc cc}
\toprule
& & \multicolumn{2}{c}{Runtime (sec.)} & \multicolumn{2}{c}{Dynamic Degree} \\
\cmidrule(lr){3-4} \cmidrule(lr){5-6}
 Model & $t_{\textrm{trans}}$ & Default & ALG & Default & ALG \\
\midrule
Wan 2.2        & 0.10 & 475 & 494  & 31.7  & 39.0 \\
Wan 2.1        & 0.20 & 476 & 527   & 28.9  & 39.4 \\
LTX-Video      & 0.10 & 58  & 59   & 15.5  & 21.5 \\
\bottomrule
\end{tabular}
\caption{
Comparison of CFG and \method on inference time and dynamism, measured using a single NVIDIA H200 GPU for video generation.
}
\label{tab:runtime_dynamic_degree}
\end{table}

\begin{table}[t]
\centering\small
\begin{tabular}{l ll}
\toprule
{Model} & Type & Source \\
\midrule
\multirow{ 2}{*}{Wan 2.2}& T2V &
\url{https://huggingface.co/Wan-AI/Wan2.2-T2V-A14B-Diffusers} \\
& I2V &
\url{https://huggingface.co/Wan-AI/Wan2.2-I2V-A14B-Diffusers} \\
\midrule
\multirow{ 2}{*}{Wan 2.1}& T2V &
\url{https://huggingface.co/Wan-AI/Wan2.1-T2V-14B} \\
& I2V &
\url{https://huggingface.co/Wan-AI/Wan2.1-I2V-14B-480P-Diffusers} \\
\midrule
LTX-Video & I2V &
\url{https://huggingface.co/Lightricks/LTX-Video} \\
\bottomrule
\end{tabular}
\caption{Models used in our experiments.}
\label{tab:app_models_source}
\end{table}

\vspace{0.05in}
\noindent
{\bf Computational cost.}~
\label{app:additional_results_speed}
Note that \method introduces additional inference cost compared to original CFG.
Specifically, CFG (Eq.~\eqref{eq:cfg_lpg}) requires a forward pass of two conditions $(\mathbf{x}_{\textrm{init}}, \mathbf{c})$ and $(\mathbf{x}_{\textrm{init}}, \varnothing)$, while \method requires additional computation of $(\mathbf{x}_{\textrm{init}}, \varnothing)$ for the first term of Eq.~\eqref{eq:cfg_lpg}.
Thus, \method introduces a tradeoff between inference cost and dynamic degree, which can be controlled by setting hyperparameter $t_{\textrm{trans}}$.
However, this overhead is marginal, as $\kappa(t)\neq 0$ for only few $t$ values (\emph{i.e.}, we only apply low-pass filter in the early steps; see Tab.~\ref{tab:app_models_modified}).
We present the running time in seconds to generate one video per model, for the default method (CFG) and our method (\method) in Tab.~\ref{tab:runtime_dynamic_degree}.
As shown, the additional cost introduced by \method is at most around 11\% (for Wan 2.1), while Dynamic Degree increases by on average 33\% (36.3\% for Wan 2.1).

\section{Experimental setup details}\label{app:experimental_details}
In this section, we provide additional details about our experiments, including additional experimental results (both qualitative and quantitative), inference setup (model checkpoints, inference parameters), and computational resources (GPU, memory). 

\subsection{Inference setup}
\vspace{0.05in}
\noindent
{\bf Model checkpoints and configuration.}~
The overview of the model checkpoints and configurations for the four models used in our experiments are presented in Tab.~\ref{tab:app_models_source} and Tab.~\ref{tab:app_models_modified}. 
For all experiments, we use the default settings from the original model provider. 
While Wan 2.2 supports multiple resolutions (480p and 720p), we found that setting the resolution to a larger size than 480p leads to a very slow inference speed, making the evaluation using three datasets very difficult.
Thus, we resorted to 480p resolution for Wan 2.2 generation.
We note that we perform all our experiments by generating 1 video either using a single NVIDIA H200 GPU or using a single NVIDIA H100 GPU (\emph{i.e.}, no multi-GPU inference was used). 

\vspace{0.05in}
\noindent
{\bf \method configuration.}~
The bottom row of Tab.~\ref{tab:app_models_modified} summarizes the hyperparameters ($t_\textrm{trans}$, $\kappa_*$) for our main experiments (Tab.~\ref{tab:main_results}). 
To determine the hyperparameters, we take 20 prompts (out of 246, randomly chosen) from the VBench evaluation set and apply a grid search to determine the best hyperparameter set.
Specifically, we search within $t_\textrm{trans}\in \{0.04, 0.1, 0.2\}$ and $\kappa_t\in\{1.6, 2.5, 4\}$. 
Note that as shown in Fig.~\ref{fig:component_trans} and Fig.~\ref{fig:component_strength}, most small $t_\textrm{trans}$ values and moderately large $\kappa_*$ values show reasonable enhancement in dynamic degree, while maintaining video quality.
Based on our exploration, the parameters detailed in Tab.~\ref{tab:app_models_modified} are those we found to yield the most advantageous increase in dynamic degree while maintaining or even often improving the overall generation quality.
For the Gaussian blur experiments of our component analysis in Sec.~\ref{sec:component_analysis}, we use kernel size of $0.05\times \textrm{height}$ pixels and $\sigma_\textrm{blur}$ of 80.
Finally, note that we low-pass filter the \emph{latents} instead of the raw input conditioning image for all our \method experiments, as the latents are the actual inputs to the denoiser model.

\begin{table}[t]
    \centering\small
    \setlength\tabcolsep{3pt}
    \begin{tabular}{@{}ll c c c c @{}}
        \toprule
        & {Model} & Wan 2.2 & Wan 2.1 & LTX-Video \\
        \midrule
        \multirow{ 6}{*}{Base config.} &
        Video length & 5s & 5s & 5s \\
        &Num. of frames & 81 & 81 & 121 \\
        &Denoising steps & 50 & 50 & 30 + 10 \\
        &Resolution & 832$\times$480 & 832$\times$480 & 1216$\times$704 \\
        &CFG scale & 5.0 & 5.0 & 3.0 \\
        &Miscellaneous & - & CLIP conditioning &  Two-stage inference \\
        \midrule
        \multirow{ 4}{*}{\method config.} &
        $t_{\textrm{trans}}$ & 0.1 & 0.2 & 0.1 \\
        &$\kappa_*$ & 2.5 & 2.5 & 4.0 \\
        & First-frame override & True & True & False \\
        & Low-pass filter delaying & 0.04 & - & - \\
        \bottomrule
    \end{tabular}
    \caption{Details for experiment with each image-to-video model. }
    \label{tab:app_models_modified}
\end{table}

\subsection{Evaluation set}\label{app:evaluation_set}
As explained in Sec.~\ref{sec:experiments}, we utilize and curate from three benchmark datasets, namely, VBench-I2V~\citep{vbench} test set, PE Video Dataset~\citep{pvd}, and VidProM~\citep{vidprom}. 
\textbf{VBench-I2V} is an image-to-video benchmark dataset, consisting of image-prompt pair data, as well as a set of evaluation metrics to assess the I2V generation performance. 
From the image-prompt pairs, we select all prompts except for those used for measuring background quality or camera motion instruction following. 
We exclude them due to the high inference cost while focusing on the motion and video quality related metrics which are the focus of our evaluation.
This gives us 246 prompts and images from the VBench-I2V dataset. 

Additionally, to evaluate the effectiveness of \method for improving motion in real video frames, we curate an image-to-video benchmark dataset from \textbf{PE Video Dataset (PVD)}. 
PVD is a video-text pair dataset including pairs of real videos and expert-annotated captions, and we take 100 random video-caption pair from the entire set. 
From each video, we took the first frame of the video and used it to construct the input image-caption pair dataset for the image-to-video generation. 

Finally, we curate 750 prompts from \textbf{VidProM} to evaluate the capability of \method to enhance image-to-video motion dynamics for synthetic image inputs generated using a state-of-the-art text-to-image (T2I) model. VidProM is a text prompt dataset for text-to-video generation, and we randomly select 750 prompts (after filtering our erroneous prompts) from the set, and use a T2I generation model (\emph{i.e.}, FLUX.1-dev~\citep{flux}) to construct image-prompt pairs for image-to-video generation. 

For all results in our experiments in Sec.~\ref{sec:experiments} and Sec.~\ref{sec:component_analysis}, we evaluate each method with one seed and using all prompts in each curated benchmark dataset. It is also worth noting that our evaluations utilize 5-second videos across all base models. 
This presents a more demanding scenario compared to the VBench leaderboard~\citep{vbench} for I2V generation, which reports results for 2-second videos, demonstrating the capability of \method to maintain performance over longer temporal sequences.

\subsection{Motion augmentation of prompts}
In this section, we explain in detail the prompt motion-augmentation technique used for the results in Tab.~\ref{tab:tab_enhanced} of Sec.~\ref{sec:component_analysis}. 
While \method enhances motion dynamics of I2V generation by adaptively modulating the frequency component during the video sampling process, it is also possible to alter the input text prompts so that the generated videos become more dynamic, orthogonally to \method. 
To test the efficacy of such prompt augmentation, as well as \method's effectiveness combined with such method, we use an LLM (Gemini 2.5 Flash; \citep{gemini}) in order to modify the prompts to become more dynamic by emphasizing motion-related parts in the prompts.
Specifically, we provided all 246 prompts from the VBench-I2V test set~\citep{vbench} and instructed Gemini 2.5 Flash to make each prompt more dynamic, by using the following prompt: ``\textit{For each of these text descriptions of video, make it more dynamic to greatly enhance motion. Do not add new elements; enhance the existing descriptions.}''      
We then provided all 246 prompts right after this instruction, with each prompt line-separated.
The LLM was instructed not to add any new elements to ensure that the prompts do not introduce objects not present in the input image, potentially introducing misalignments between images and prompts.
We include all prompts resulting from this augmentation in our supplementary materials.
Then, the baseline method (CFG) and our method (\method) were tested using this motion-enhanced VBench test set on Wan 2.2, as presented in Tab.~\ref{tab:tab_enhanced}. 
It is worth noting that both methods benefit from the prompt augmentation, but \method without augmentation already surpasses CFG in terms of dynamism.

\subsection{Input prompts for generation}
Table~\ref{tab:input_prompts} summarizes the text prompts and specific image-to-video models utilized to generate the videos shown in Fig.~\ref{fig:main_qual}.

\begin{table}[htbp]
    \centering
    \footnotesize
    \begin{tabularx}{\textwidth}{@{}l l X@{}}
        \toprule
        \textbf{Location} & \textbf{Model} & \textbf{Text Prompt} \\
        \midrule
        \multicolumn{3}{@{}l}{\textit{\textbf{Top (Larger Videos)}}} \\
        \midrule
        Video 1 & Wan 2.2 & A white car is swiftly driving on a dirt road near a bush, kicking up dust. \\
        Video 2 & Wan 2.2 & A beach ball floats up into the sky, realistic handheld style, style of Ken Loach, camera pans up following. \\
        Video 3 & LTX-Video & A dog leaping through the air to catch a frisbee in a sunny park. \\
        Video 4 & Wan 2.2 & create an image depicting the moment Luffy traverses through the dimensional rift, with vibrant colors and swirling energies 8k [the shot should create a lasting impression of horror and shock]. \\
        \midrule
        \multicolumn{3}{@{}l}{\textit{\textbf{Bottom (Smaller Videos - Clockwise from top-left)}}} \\
        \midrule
        Video 1 & Wan 2.2 & A space station orbited above the Earth. \\
        Video 2 & Wan 2.2 & A princess dancing in a pink dress under a pink and purple sky, tiny shining particals falling from the sky. \\
        Video 3 & Wan 2.1 & High Speed Super Cute Nuclear Christmas Explossiono, Sci-fi, Virtual Reality 3D. \\
        Video 4 & Wan 2.1 & Create an 8k video of a beautiful woman with long blonde hair and blue eyes. She is wearing a red dress and high heels. She is walking on a busy street in New York City, smiling and waving at the camera. The video should have a smooth zoom in and out effect, and a slow motion effect when she turns her head. \\
        Video 5 & Wan 2.2 & A child kicks a bear in a Finnish forest. the weather is sunny. zoom in. \\
        Video 6 & Wan 2.1 & A photorealistic blue haired Tooth Fairy called Molly fights the many teeth creatures with her magical toothbrush. \\
        Video 7 & Wan 2.1 & A beautiful girl with dark brown hair jumps out of the sparks of fire, The extension of gold thread, the splash of gold thread, the diffusion of ink, the smearing of ink. \\
        Video 8 & Wan 2.2 & A large pot of soup filled with vegetables and meat. \\
        \bottomrule
    \end{tabularx}
    \caption{Input text prompts and models corresponding to the videos in Fig.~\ref{fig:main_qual}.}
    \label{tab:input_prompts}
\end{table}

\subsection{Evaluation metrics}
In this section, we provide a detailed explanation for the definition of each metric used in all our evaluation results (Sec.~\ref{sec:experiments} and Sec.~\ref{sec:component_analysis}), including the VBench metrics~\citep{vbench}, DOVER~\citep{DOVER}, and VisionReward~\citep{visionreward}.

\vspace{0.05in}
\noindent
{\bf VBench: Motion-related metrics.}~
VBench includes one metric that assesses the degree of motion presented in the generated videos (\emph{Dynamic Degree}). Note that the \textit{VBench-Avg.} metric reported in Sec.~\ref{sec:experiments} is the average value of all VBench metrics explained in this section, including Dynamic Degree.

\begin{itemize}[leftmargin=*,itemsep=0.5mm]
    \item \emph{Dynamic Degree}: This is the metric that is central to our evaluation of the enhanced motion dynamics of videos. For a single video, Dynamic Degree is computed by computing the magnitude of the top-5\% optical flow between frames using RAFT~\citep{raft}, and then thresholding this value to determine whether each frame interval is \emph{dynamic} or \emph{static}. Then, the video is labeled as ``dynamic'' if the percentage of the dynamic interval exceeds a certain threshold. Both thresholds (frame interval flow magnitude, percentage of dynamic intervals) are determined adaptively according to the video resolution in order to ensure a fair cross-resolution comparison. Additionally, the FPS (frame per second) values is normalized to 8 FPS.
\end{itemize}

\vspace{0.05in}
\noindent
{\bf VBench: Image-to-video consistency metrics.}~
We employ the I2V Subject Consistency metric from the VBench evaluation suite. This metric assess the fidelity of the video compared to the given input conditioning image. Note that the reported \textit{VBench-I2V} metric in Sec.~\ref{sec:experiments} refers to I2V Subject Consistency.

\begin{itemize}[leftmargin=*,itemsep=0.5mm]
    \item \emph{I2V Subject Consistency}: This metric assesses the consistency of the subject in the input image and the subject in the generated video frames. It is computed by measuring the DINO~\citep{dino} similarity between input image and all video frames. Additionally, DINO similarity between consecutive frames is measured, and take the weighted average of these two similarities is used as the final metric value.
\end{itemize}

\vspace{0.05in}
\noindent
{\bf VBench: Video quality metrics.}~
Video quality metrics include 5 sub-metrics: \emph{Subject Consistency}, \emph{Temporal Flickering}, \emph{Motion Smoothness}, \emph{Aesthetic Quality}, and \emph{Imaging Quality}. The first 3 metrics assess the quality of the video in a temporally dependent manner, and the last 2 metrics measure the frame-wise quality. Note that our reported average metric \textit{VBench-QS} in Sec.~\ref{sec:experiments} is the average of these five metrics.

\begin{itemize}[leftmargin=*,itemsep=0.5mm]
    \item \emph{Temporal Quality - Subject Consistency}: This measures the consistency of the subject within a video, and is calculated by computing the DINO feature similarity between frames.
    \item \emph{Temporal Quality - Temporal Flickering}: Unlike the first two consistency metrics (Subject Consistency and Background Consistency), which gauge semantic consistency, this metric focuses on the consistency of high-frequency local details by computing the mean absolute difference of frames.
    \item \emph{Temporal Quality - Motion Smoothness}: This metric assesses the smoothness of the generated motions using the motion priors in a video frame interpolation model~\citep{amt}.
    \item \emph{Frame-wise Quality - Aesthetic Quality}: This evaluates how aesthetically beautiful the individual frames are, using the LAION aesthetic predictor~\citep{aesthetic}. This predictor takes into account various beauty aspects including color combination, lighting, photo-realism, and the layout of the image.
    \item \emph{Frame-wise Quality - Imaging Quality}: Measures how distortion-free the frames are. Distortion includes various imaging-related factors, such as over-exposure, noise, and blur, and is measured using the MUSIQ~\citep{musiq} image quality predictor.
\end{itemize}

\vspace{0.05in}
\noindent
{\bf DOVER: Aesthetic and technical quality metrics.}~
DOVER~\citep{DOVER} evaluates video quality by separating two perceptual dimensions: \emph{aesthetic} (semantic appeal, composition, meaningfulness) and \emph{technical} (low-level distortions), where each corresponds to one branch of the model. The overall DOVER score reported in Sec.~\ref{sec:experiments} is a weighted sum of the two, using weights defined in the original DOVER work~\citep{DOVER}, determined to best match human judgments.

\begin{itemize}[leftmargin=*,itemsep=0.5mm]
    \item \emph{DOVER – Aesthetic Quality}:  
    This metric measures the aesthetic perception of a video, focusing on semantic content, composition, object arrangement, and overall visual pleasantness. This is computed from an \emph{Aesthetic View} produced by spatial downsampling and sparse temporal sampling, which preserve semantics while suppressing distortions.
    \item \emph{DOVER – Technical Quality}:  
    This measures the presence of low-level distortions, such as blur, noise, exposure errors, and jitter. DOVER constructs a \emph{Technical View} by sampling fragmentary spatiotemporal patches so that the metric reflects technical fidelity only, not global semantics.
\end{itemize}

\vspace{0.05in}
\noindent
{\bf VisionReward: Human-preference reward model for videos and images.}~
VisionReward~\citep{visionreward} is a multi-dimensional human preference reward model for images and videos, built using a VLM model (QWen2-VL~\citep{qwen}). It decomposes human judgments into several dimensions (e.g., alignment, composition, stability, dynamics) via binary checklists; these are linearly weighted and summed into a single interpretable score. We report the single overall VisionReward score in Sec.~\ref{sec:experiments}. 

\begin{itemize}[leftmargin=*,itemsep=0.5mm]
\item \emph{VisionReward - Overall score}: Videos are evaluated on multiple dimensions (\emph{e.g.}, motion realism, camera motion, stability, physics) via checklist questions; answers are mapped to binary features and a linear regression produces the final score. 

\end{itemize}

\section{Additional results}\label{app:additional_results}
{
In this section, we present and explain additional experimental results of \method, as well as the information regarding the experimental results demonstrated in our main text. As explained in Fig.~\ref{fig:main_qual}, all prompts and models used for generation of videos, including Fig.~\ref{fig:main_qual}, can be found in our supplementary materials.
}

\subsection{Qualitative examples}\label{app:additional_results_qualitative}
{
We first provide additional qualitative examples for \method in the following sections as explained below. The qualitative example videos can be viewed in a playable video file format in our supplementary materials.

\begin{itemize}[leftmargin=*,itemsep=0mm]
    \item Appendix~\ref{app:qual}: Additional qualitative results of video generation.
    \item Appendix~\ref{app:design_choice}: Visualization of the effects of the design choice for the first unconditional term in \method on the visual quality of the generated video quality, which is discussed in Sec.~\ref{sec:method}. (Fig.~\ref{fig:uncond_term_design_choice})

\end{itemize}
}

\subsubsection{Additional qualitative results}\label{app:qual}
We provide additional qualitative examples for our experiments in Fig~\ref{fig:app_qual_1}. More qualitative example videos can be seen in video format in our supplementary materials, along with their prompts and I2V models used for generation.

\begin{figure*}[t]
  \centering
  \includegraphics[width=.98\linewidth]{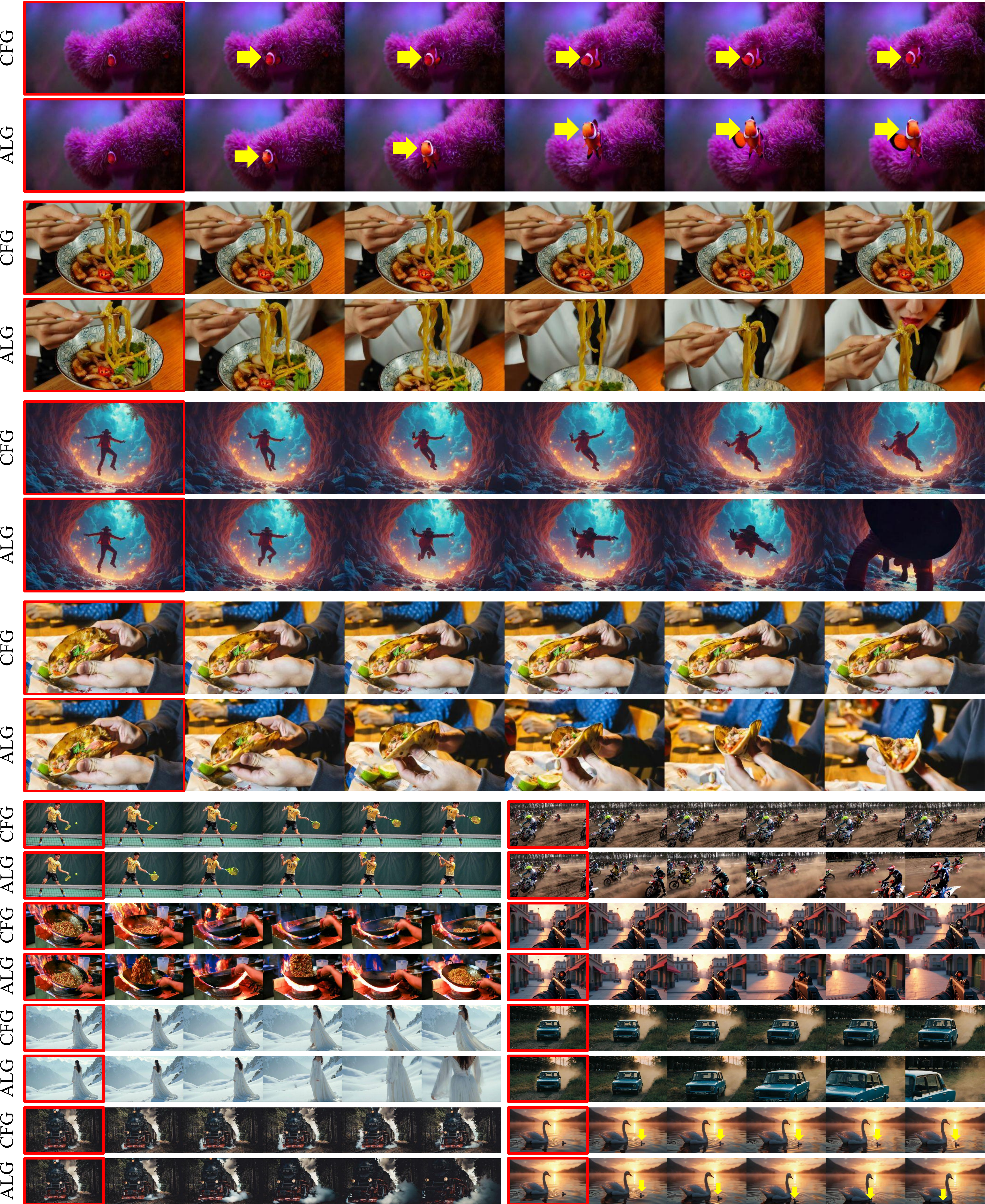}
  \vspace{-5pt}
  \caption{
  {\bf Additional qualitative comparison between \method and CFG.}~ 
  We compare the videos generated by using default image-to-video generation method (CFG) and our method (\method).
  The input conditioning frames are denoted with red outline.
  We observe that the videos using \method show more dynamic motion (\emph{e.g.}, larger object movement, animal movement, or human action, and more complex background movements).
  The list of prompts and models used for each video is included in the supplementary material.
  }
  \label{fig:app_qual_1}
  \vspace{-15pt}
\end{figure*}

\newpage
\clearpage

\subsubsection{Qualitative comparison for the design choice of \method}\label{app:design_choice}
{
We visualize the qualitative differences that arise when using the low-pass filtered latent for all unconditional terms in Eq.~\eqref{eq:cfg_lpg} (see Sec.~\ref{sec:method} for more details) in Fig.~\ref{fig:uncond_term_design_choice}.
As shown, using the low-pass filtered latent for all unconditional terms often result in unstable video generation results, often characterized by distorted video frames or abrupt changes of scenes. 
}
\vspace*{\fill}
\begin{center}
    \begin{figure}[h]
  \begin{subfigure}[h]{\textwidth}
    \centering
    \includegraphics[width=\textwidth]{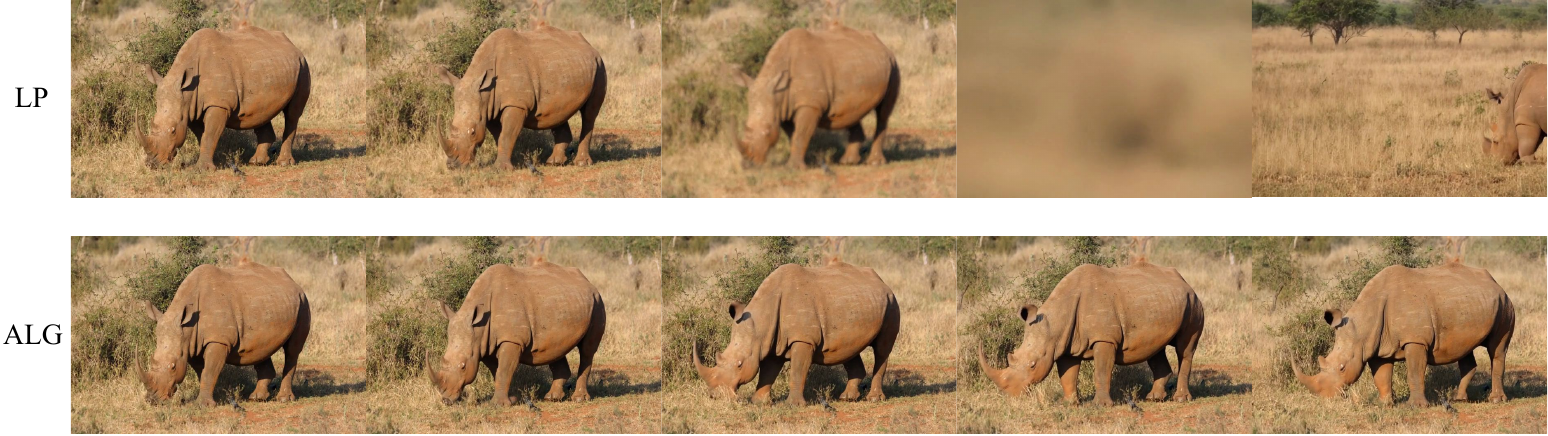}
    \caption{Distortion of video frames}
    \label{fig:uncond_choice_1}
  \end{subfigure}
  \vfill
  \vspace{5pt}
  \begin{subfigure}[h]{\textwidth}
    \centering
    \includegraphics[width=\textwidth]{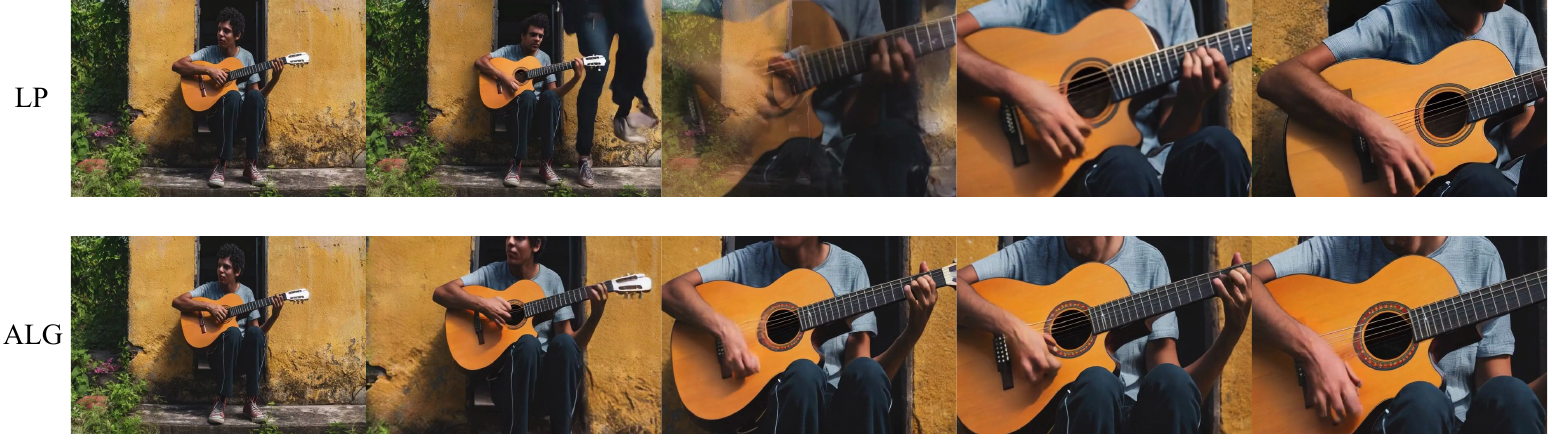}
    \caption{Abrupt scene change}
    \label{fig:uncond_choice_2}
  \end{subfigure}
  \caption{\textbf{Visual examples that warrant the design choice of \method.} Using low-pass filtered input image for all terms of classifier-free guidance (denoted LP) often results in (a) distorted video frames, or (b) abrupt scene changes. \method avoids this issue by grounding the generation in the original image's precise details and simultaneously providing motion guidance from the filtered image to enhance motion. This ensures both stability and visual integrity (see Sec.~\ref{sec:method}) while enhancing video dynamism.}
  \label{fig:uncond_term_design_choice}
\end{figure}
\end{center}
\vspace*{\fill}

\newpage
\clearpage

\subsection{Evaluation results}\label{app:additional_results_eval_reults}
In this section, we present additional quantitative experimental results.

\subsubsection{Full experimental results}
\begin{table}[t]
    \centering\footnotesize
    \vspace{2pt}
    \setlength\tabcolsep{4.6pt}
    \begin{tabular}{lcc>{\columncolor[HTML]{DAE7ED}}ccccccccc|ccc|c}
    \toprule
    & & & {Dynamic} & \multicolumn{8}{c|}{VBench} & \multicolumn{3}{c|}{DOVER} & \multicolumn{1}{c}{Vision}\\
    Benchmark & Model & Method &
    Degree & Avg. & QS & I2V & TF & AQ & SC & IQ & MS &
    Overall & Aes. & Tech. & {Reward} \\
        \midrule
        \multirow{7.5}{*}{VBench~\citep{vbench}}
        & \multirow{2}{*}{Wan 2.2}
        & CFG & 31.7 & 79.6 & \textbf{85.4} & 98.5 & \textbf{98.2} & 63.2 & \textbf{96.2} & 69.9 & \textbf{98.9} & 0.635 & 0.768 & 0.509 & \textbf{0.183} \\
        & & \method (Ours) & \textbf{39.0} & \textbf{80.5} & 85.2 & \textbf{98.5} & 97.8 & \textbf{63.5} & 95.8 & \textbf{70.0} & 98.7 & \textbf{0.637} & \textbf{0.781} & \textbf{0.530} & 0.182 \\
        \cmidrule(lr){2-16}
        & \multirow{2}{*}{Wan 2.1}
        & CFG & 28.9 & 79.1 & \textbf{85.3} & \textbf{98.3} & \textbf{98.2} & \textbf{63.2} & \textbf{96.2} & \textbf{69.9} & \textbf{98.9} & \textbf{0.618} & \textbf{0.767} & \textbf{0.509} & \textbf{0.179} \\
        & & \method (Ours) & \textbf{39.4} & \textbf{80.0} & 84.5 & 98.0 & 98.0 & 62.4 & 95.0 & 68.3 & 98.8 & 0.614 & 0.761 & 0.508 & 0.176 \\
        \cmidrule(lr){2-16}
        & \multirow{2}{*}{\makecell{LTX-Video}}
        & CFG & 15.5 & 77.8 & \textbf{85.9} & \textbf{99.1} & \textbf{99.4} & \textbf{62.4} & \textbf{98.2} & \textbf{70.0} & \textbf{99.6} & 0.625 & 0.755 & \textbf{0.527} & 0.175 \\
        & & \method (Ours) & \textbf{21.5} & \textbf{78.2} & 85.4 & 98.9 & 99.2 & 61.6 & 97.1 & 69.6 & 99.6 & \textbf{0.626} & \textbf{0.765} & 0.522 & \textbf{0.175} \\
        \midrule
        \multirow{7.5}{*}{PVD~\citep{pvd}}
        & \multirow{2}{*}{Wan 2.2}
        & CFG & 65.0 & 79.4 & 79.6 & 94.2 & \textbf{97.2} & 49.5 & \textbf{89.7} & 62.0 & \textbf{98.4} & 0.484 & 0.631 & 0.389 & 0.145 \\
        & & \method (Ours) & \textbf{69.0} & \textbf{80.3} & \textbf{79.6} & \textbf{95.0} & 96.8 & \textbf{50.0} & 89.7 & \textbf{63.0} & 98.3 & \textbf{0.512} & \textbf{0.660} & \textbf{0.415} & \textbf{0.145} \\
        \cmidrule(lr){2-16}
        & \multirow{2}{*}{Wan 2.1}
        & CFG & 66.0 & 79.4 & \textbf{79.2} & 94.0 & \textbf{97.3} & \textbf{49.7} & \textbf{88.3} & 62.2 & \textbf{98.4} & 0.492 & 0.645 & 0.393 & \textbf{0.145} \\
        & & \method (Ours) & \textbf{74.0} & \textbf{80.3} & 78.8 & \textbf{94.2} & 97.0 & 49.4 & 86.1 & \textbf{63.3} & 98.2 & \textbf{0.529} & \textbf{0.676} & \textbf{0.428} & 0.141 \\
        \cmidrule(lr){2-16}
        & \multirow{2}{*}{\makecell{LTX-Video}}
        & CFG & 66.7 & 80.0 & \textbf{79.4} & \textbf{96.2} & \textbf{98.6} & 49.4 & \textbf{91.4} & 58.5 & \textbf{99.4} & 0.428 & 0.490 & 0.391 & \textbf{0.123} \\
        & & \method (Ours) & \textbf{77.0} & \textbf{81.3} & 79.3 & 95.3 & 98.2 & \textbf{49.9} & 89.0 & \textbf{60.3} & 99.2 & \textbf{0.446} & \textbf{0.533} & \textbf{0.393} & 0.122 \\
        \midrule
        \multirow{7.5}{*}{VidProM~\citep{vidprom}}
        & \multirow{2}{*}{Wan 2.2}
        & CFG & 27.3 & 79.1 & 85.6 & \textbf{98.2} & \textbf{98.7} & \textbf{67.9} & \textbf{96.3} & \textbf{66.2} & \textbf{99.2} & \textbf{0.658} & \textbf{0.787} & 0.560 & 0.096 \\
        & & \method (Ours) & \textbf{30.5} & \textbf{79.5} & \textbf{85.6} & 98.0 & \textbf{98.7} & 67.7 & 96.2 & 66.1 & 99.2 & 0.657 & 0.785 & \textbf{0.560} & \textbf{0.096} \\
        \cmidrule(lr){2-16}
        & \multirow{2}{*}{Wan 2.1}
        & CFG & 27.8 & 79.1 & \textbf{85.6} & \textbf{98.1} & \textbf{98.8} & \textbf{67.6} & \textbf{96.2} & \textbf{66.0} & \textbf{99.3} & \textbf{0.652} & \textbf{0.776} & \textbf{0.558} & \textbf{0.093} \\
        & & \method (Ours) & \textbf{36.3} & \textbf{80.0} & 85.1 & 97.8 & 98.7 & 67.0 & 95.4 & 65.3 & 99.2 & 0.648 & 0.774 & 0.554 & 0.092 \\
        \cmidrule(lr){2-16}
        & \multirow{2}{*}{\makecell{LTX-Video}}
        & CFG & 18.2 & 78.1 & \textbf{85.9} & \textbf{99.2} & 99.5 & \textbf{67.6} & \textbf{97.8} & \textbf{64.7} & 99.6 & \textbf{0.627} & \textbf{0.764} & \textbf{0.527} & \textbf{0.089} \\
        & & \method (Ours) & \textbf{23.0} & \textbf{78.7} & 85.7 & 99.1 & \textbf{99.5} & 67.6 & 97.5 & 64.5 & \textbf{99.6} & 0.625 & 0.763 & 0.523 & 0.088 \\
        \midrule
    \end{tabular}
    \caption{{\bf Comparison of CFG and \method across all three models and benchmark datasets}. \method consistently improves Dynamic Degree while maintaining video quality (all metrics except Dynamic Degree), leading to higher VBench average score (VBench-Avg.) in all cases.
    }
    \label{tab:app_full}
\end{table}
We report the full experimental results for all three models (Wan 2.1/2.2~\citep{wan}, LTX-Video~\citep{ltx}), and for all three benchmark datasets (VBench~\citep{vbench}, PVD~\citep{pvd}, VidProM~\citep{vidprom}) in Table~\ref{tab:app_full}. 
Note that in this result, we show individual scores of each metric suite (VBench and DOVER). Specifically, for VBench, we report Temporal Flickering (TF), Aesthetic Quality (AQ), Subject Consistency (SC), Imaging Quality (IQ), and Motion Smoothness (MS) alongside the aggregate scores. For DOVER, we provide the breakdown into Aesthetic (Aes.) and Technical (Tech.) metric scores in addition to the Overall score.
Consistent with the results reported in Sec.~\ref{sec:experiments} and Sec.~\ref{sec:component_analysis}, \method improves Dynamic Degree across all models and benchmarks while maintaining quality metrics. 
As a result, VBench average score (VBench-Avg.) increases in \method for all cases in Tab.~\ref{tab:app_full}.

\subsubsection{Applying low-pass filter to the input image with CogVideoX}\label{app:additional_results_suppression}
\begin{wrapfigure}{r}{0.325\textwidth}
  \vspace{-13pt}
  \includegraphics[width=\linewidth]{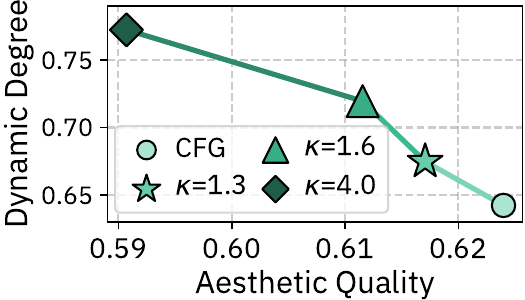}
  \caption{Low-pass filtering input image enhances motion in CogVideoX.}
  \label{fig:app_suppression_restoration_cogvideox}
\end{wrapfigure}

In order to show that the dynamism gap between text-to-video and image-to-video model variants shown in Sec.~\ref{sec:challenges} is not a phenomenon limited to Wan~\citep{wan} models, we report the diagnostic results with an additional open-source video model that has both text-to-video and image-to-video model variation, namely, CogVideoX~\citep{cogvideox}. 
We show the enhancement of dynamic motion upon applying low-pass filtering with varying strengths, similarly to Fig.~\ref{fig:t2v_to_i2v_suppression_restoration_a} (Wan 2.1 results).
The results are visualized in Fig.~\ref{fig:app_suppression_restoration_cogvideox}. 
As shown, we observe that stronger low-pass filtering results in enhanced dynamic degree of the generated videos and a loss of per-frame video quality (aesthetic quality).
This finding with CogVideoX is aligned with the results with Wan 2.1 presented in Fig.~\ref{fig:t2v_to_i2v_suppression_restoration_a} of Sec.~\ref{sec:challenges}, and further supports our claim that low-pass filtering the input image results mitigates the motion suppression effect in I2V models.

\newpage
\clearpage

\subsection{Feature map visualization}\label{app:additional_results_fmap}
We provide additional results of feature map visualization as shown in Fig.~\ref{fig:shortcut} of Sec.~\ref{sec:challenges}.
For Fig.~\ref{fig:shortcut}, similarly to DINOv2~\citep{dinov2} and REPA~\citep{repa}, we inspect the middle layers of the DiT denoiser of Wan 2.1 by selecting the 5th frame of the intermediate activation at this layer.
We provide additional visualizations for more diverse prompts, DiT layers, and $t$ values in Fig.~\ref{fig:app_shortcut_l14} and Fig.~\ref{fig:app_shortcut_l21}.
Additionally, we include the feature map visualization for our method (\method), which exhibits a similar behavior to the mitigation of the shortcut effect seen in na\"ive low-pass filtering, as \method applies low-pass filtering at the early stages of the denoising process (where shortcut effect occurs predominantly).

\vspace*{\fill}
\begin{center}
    \begin{figure}[h]
  \centering
  \includegraphics[width=\linewidth]{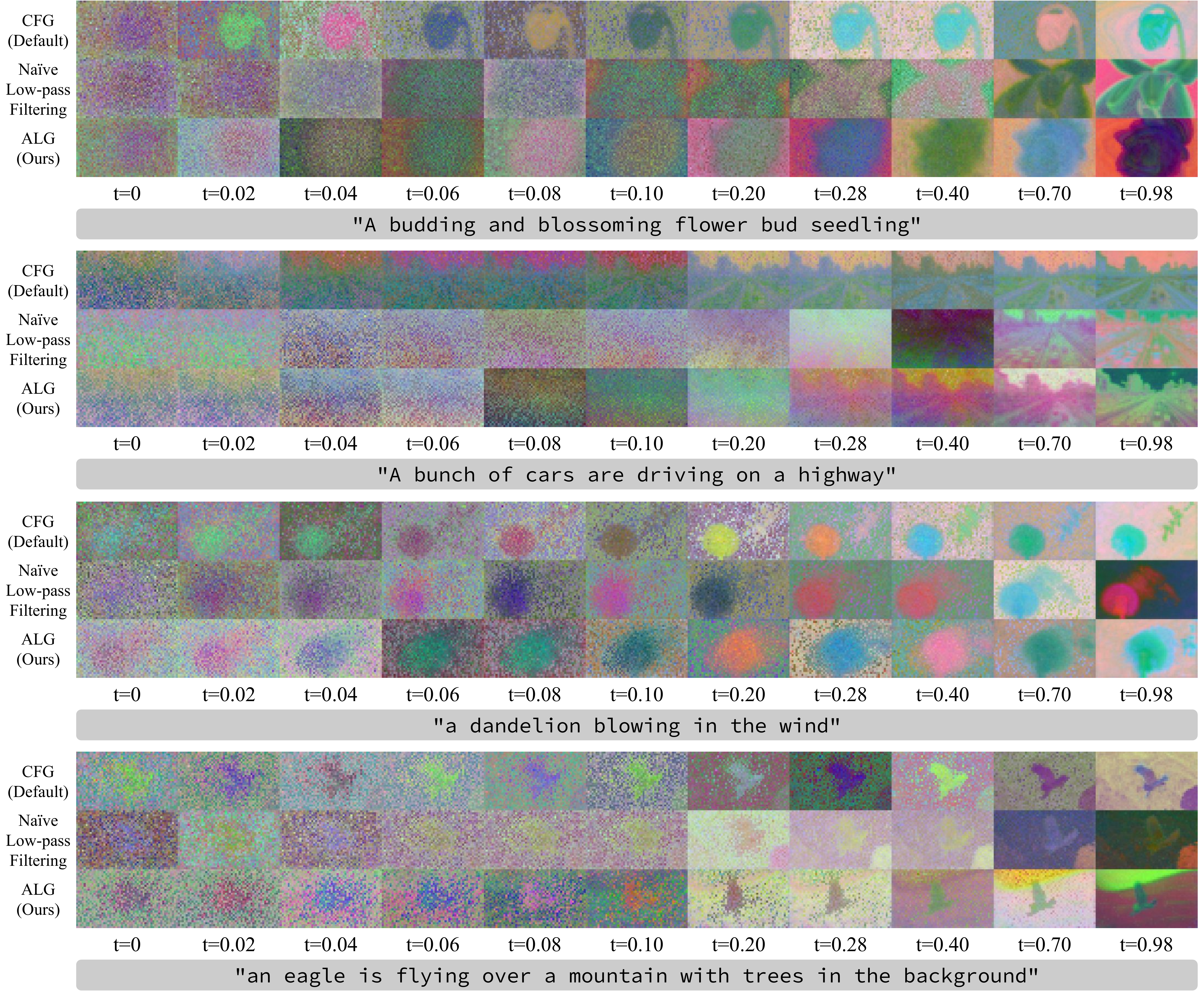}  
  \caption{\textbf{
  Visualization of shortcut effect in I2V generation for the 15th layer of the DiT backbone.}
  For all default video generation results, we observe a premature refinement of the feature maps similar to Fig.~\ref{fig:shortcut}.
  Low-pass filtering the input image avoids the shortcut effect and get refined more gradually.
  We observe similar effects in the case of our method (ALG), as it applies low-pass filter in the early stages of the sampling process.
  Best viewed in zoomed and colored monitor.}
  \label{fig:app_shortcut_l14}
\end{figure}
\end{center}
\vspace*{\fill}

\begin{figure}[ht]
  \centering
  \includegraphics[width=\linewidth]{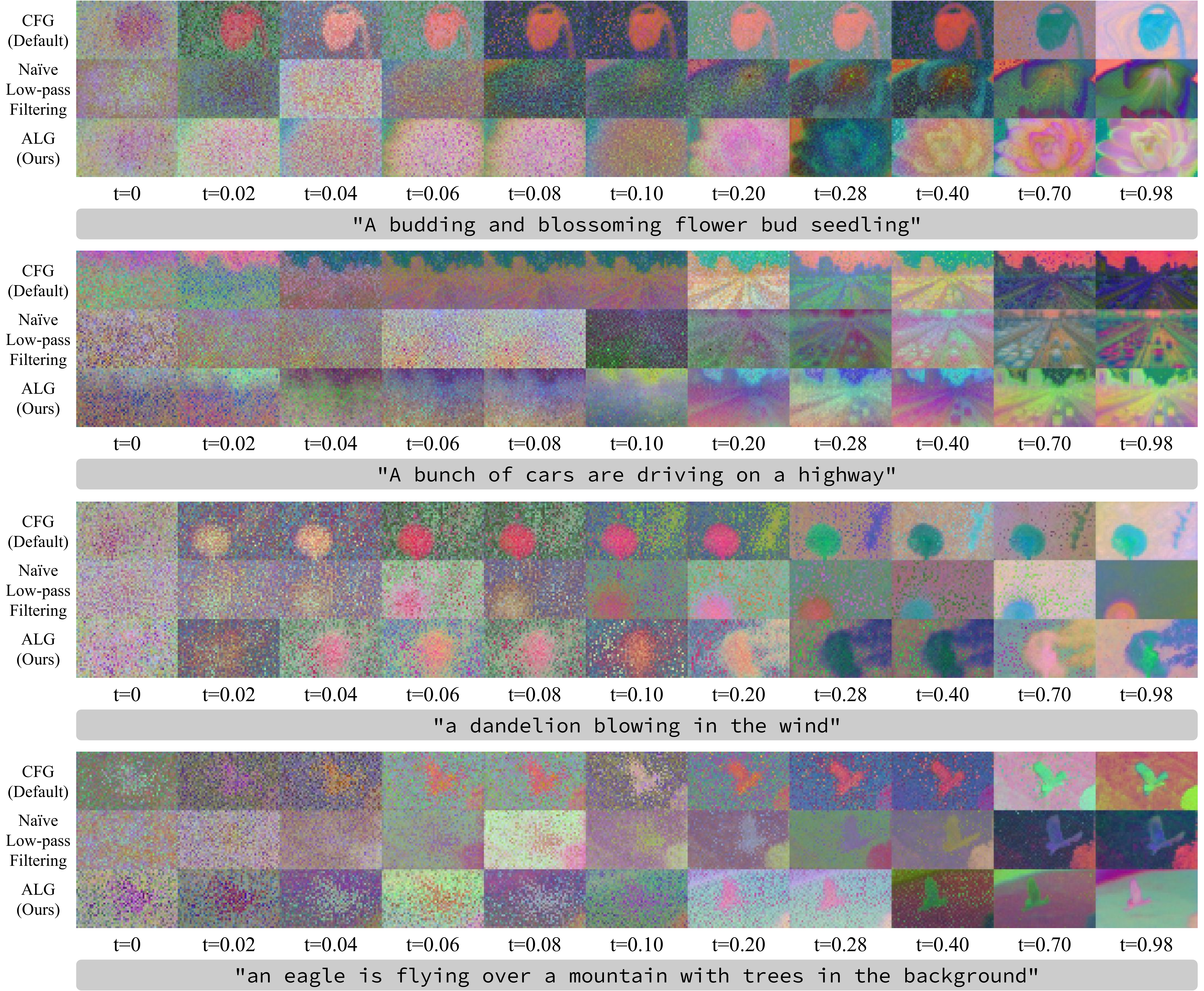}  
  \caption{\textbf{
  Visualization of shortcut effect in I2V generation for the 22th layer of the DiT backbone.}
  Similar to Fig.~\ref{fig:app_shortcut_l14}, we observe that baseline method suffers a premature refinement of feature maps while low-pass filtering mitigates this effect, resulting in a more gradual refinement.
  Additionally, we observe similar mitigation in the case of our method (ALG), as it applies low-pass filter in the early stages of the sampling process.
  Best viewed in zoomed and colored monitor.}
  \label{fig:app_shortcut_l21}
\end{figure}
}
\end{document}